\documentclass[12pt, helvetica]{article}
\usepackage{amsmath}
\usepackage{graphicx}
\usepackage{enumerate}
\usepackage{dsfont}
\usepackage{amsmath,bm}
\usepackage{amsmath}
\usepackage{amssymb}
\usepackage{amsthm}
\usepackage{mathtools}
\usepackage{amsbsy}
\usepackage{multirow}
\usepackage{authblk}
\usepackage{enumitem}
\usepackage{nameref}
\usepackage{xcolor}
\usepackage{hyperref}
\usepackage[numbers]{natbib}
\usepackage{bm}
\usepackage{tikz}
\usepackage{authblk}
\usepackage{amsfonts}
\usepackage{placeins}
\usepackage{url}
\usetikzlibrary{positioning}
\usepackage[english]{babel}
\usepackage[T1]{fontenc}
\usepackage{todonotes}
\usepackage{wrapfig}

\newcommand{\blind}{0}

\theoremstyle{remark}

\begin{document}

\def\spacingset#1{\renewcommand{\baselinestretch}%
{#1}\small\normalsize} \spacingset{1}

%%%%%%%%%%%%%%%%%%%%%%%%%%%%%%%%%%%%%%%%%%%%%%%%%%%%%%%%%%%%%%%%%%%%%%%%%%%%%%

\if0\blind
{
  \title{\bf Gaussian Process Decorrelation for Spatiotemporal Deep Learning-Based Snow Water Equivalent Prediction}
  \author{Colin Fenster$^1$, Adrienne Marshall$^2$,  Soutir Bandyopadhyay$^1$, Daniel McKenzie$^1$\\
    \footnotesize$^1$Department of Applied Mathematics and Statistics, Colorado School of Mines, \\ 
    \footnotesize$^2$Hydrologic Science and Engineering Program, Colorado School of Mines }
  \date{}
  \maketitle
} \fi

\if1\blind
{
  \bigskip
  \bigskip
  \bigskip
  \begin{center}
    {\LARGE\bf Title}
\end{center}
  \medskip
} \fi

\bigskip
\begin{abstract}
%{\color{blue} [SB: We will return to the abstract after completing the rest of the paper.]} 
In the Western United States, snowmelt is essential to the agricultural industry in addition to being a key source of municipal drinking water. Consequently, accurate  snowpack forecasting is critical for water policy and management. Automated Snow Telemetry (SNOTEL) stations provide accurate daily measurements of snow water equivalent (SWE) that exhibit strong correlations in space and in time. We tackle the problem of predicting future SWE values across the SNOTEL network.

Specifically, we use a Gaussian Process-based linear transformation to remove spatial correlations before training a long short-term memory (LSTM) neural network on the decorrelated SWE data. This approach allows the LSTM to learn a clean temporal signal at each station. We show that this separation of spatial and temporal components yields better predictive success than multiple baseline models. 

Furthermore, we incorporate conformal prediction to quantify uncertainty in the resulting SWE forecasts, providing a distribution-free approach to illustrate a potential framework for establishing predictive intervals for spatiotemporal data. Together, accurate point forecasts and distribution-free uncertainty quantification provide a framework for SWE accumulation forecasting on subseasonal scales or projecting SWE with future data while motivating and supporting future work in predicting a large-scale, spatiotemporally complete SWE map.
\end{abstract}

\noindent%

\spacingset{1.45}
%\newpage
\section{Introduction}
\label{sec:intro}

Mountain snowpack is a major driver of water resources availability for temperate regions globally \citet{viviroli_mountains_2007}. Snow stores water that falls in the cool season, enhancing water availability by forming a natural reservoir \citet{barnett_potential_2005}. This snow storage is essential to supporting water demands for municipal use, agriculture \citet{qin_agricultural_2020}, and hydropower \citet{marshall_impacts_2022, cite-hydropower}. In the western U.S., about half of total runoff originates from snow \citet{li_how_2017}. Snow is also important beyond its role for water resources: it supports recreation economies \citet{jenkins_spring_2023, wobus_projected_2017} and plays an important cooling role in climate feedbacks \citet{thackeray_snow_2019}. Many ecosystems \citet{slatyer_ecological_2022} and wildlife species \citet{vega_detection_2025, barsugli_projections_2020} depend on snow. Snow sustains wetter soil moisture during the dry season \citet{harpold_sensitivity_2015}, and low snow years are associated with higher fire activity \citet{westerling_warming_2006}. Ongoing climate change is driving declines in western U.S. snow \citet{gottlieb_evidence_2024, mote_dramatic_2018} as well as episodic “snow drought” years \citet{marshall_2026_2026}, challenging the viability of this essential resource.

 %Additionally, SWE is considered a key indicator of climate change, as warming reduces the amount of snowpack present in a given year \citep{cite-mote}. Changes in SWE affect tourism and recreation, the agricultural industry, and native ecosystems \citep{cite-wolverines,cite-epa}. Furthermore, flood risk can increase dramatically during rain-on-snow events, such as the 2017 Oroville Dam crisis in California, highlighting the importance of understanding seasonal trends in snowpack and SWE for mitigating natural hazards \citep{cite-ROS}.

Given the importance of snow, accurate measurement and prediction are essential. The United States Natural Resource Conservation Service (NRCS), through the National Water and Climate Center (NWCC), maintains over 800 automated data collection sites known as Snow Telemetry (SNOTEL) stations across the Western United States \citep{cite-SNOTELprods}. These stations continuously measure the quantity of water contained in the snow, known as snow water equivalent (SWE), and associated meteorological variables. Snow accumulation primarily occurs between October 1 and June 30; we refer to this interval as the \textit{snow year}. Data from SNOTEL stations spanning the 2001--2019 snow years therefore form a temporally episodic and spatially correlated dataset for developing and evaluating SWE reanalysis models. %Of note, we remove stations with missing data, thus leaving 323 stations for analysis. 

\begin{figure}
    \centering
    \includegraphics[height = 3in, width = 4in]{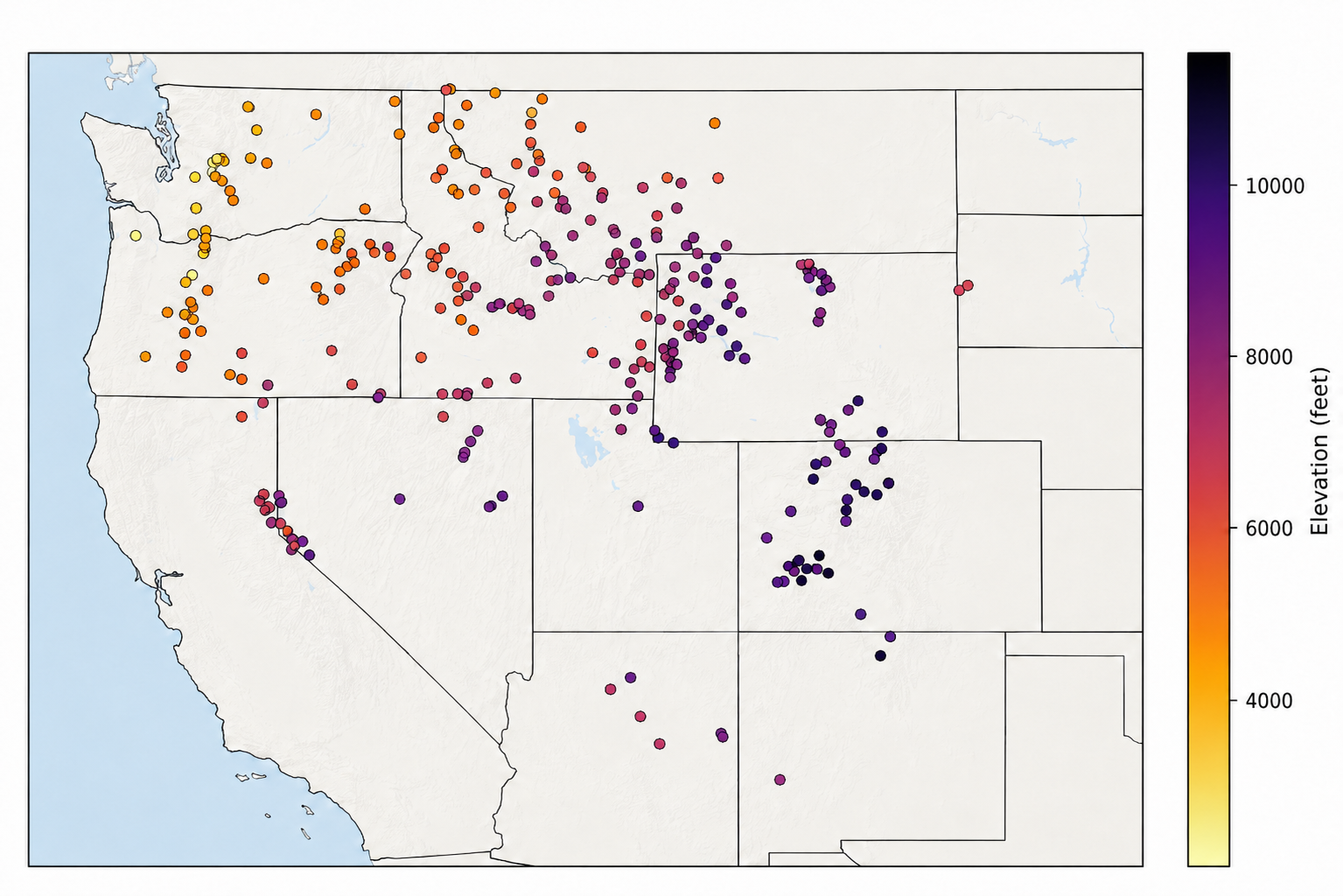}
    \caption{Spatial Map Showing 323 SNOTEL Locations Shaded by Elevation}
    \label{fig:snotel_locs}
\end{figure}

SWE prediction presents a fundamental modeling challenge because observations exhibit strong spatial and temporal correlations. While statistical approaches such as Gaussian process (GP) models explicitly account for spatial dependence, modern deep learning (DL) methods often attempt to learn spatial and temporal structure simultaneously (e.g., \citet{ cite-Thapa}). %This increases model complexity and can obscure temporal signals that are critical for accurate forecasting. Recent work has explored a variety of strategies for incorporating spatial information into machine learning models, including GP-based neural networks and spatiotemporal transformer architectures \citet{cite-zhandatta}. However, these approaches generally incorporate spatial dependence directly within the forecasting model rather than separating the spatial and temporal components of the learning problem.
%Existing DL approaches generally require spatial and temporal dependence to be modeled simultaneously. 
Here, we instead separate these two sources of dependence by applying a GP-based decorrelation transformation prior to model training. Specifically, we extend the spatial whitening framework proposed by \citet{cite-heaton} to spatiotemporal data and use a recurrent neural network \citet{werbos1990backpropagation} with long short-term memory (henceforth: an LSTM) \citet{cite-LSTM} for temporal prediction. To quantify predictive uncertainty, we construct conformal prediction intervals with distribution-free coverage guarantees \citet{lei2018distributionfree}. Summarizing, we propose a three-stage modeling pipeline combining:

\begin{enumerate}
    \item \textbf{Spatial whitening} to approximately remove spatial correlations in response and predictor variables. The model uses these spatially whitened quantities during training and subsequently applies the inverse transformation to recover predictions on the original physical SWE scale.
    \item \textbf{Temporal prediction} using an LSTM to produce sequence-to-sequence SWE predictions over the desired forecast period.
    \item \textbf{Conformal calibration} applied to the LSTM predictions in the original SWE scale, to construct prediction intervals around each point prediction. %This final stage supplements the spatial-temporal point forecasts with an empirical measure of predictive uncertainty.
\end{enumerate}

%Together, these stages define the complete prediction pipeline: the whitening transformation accounts for spatial structure, the LSTM models temporal evolution, and conformal prediction quantifies uncertainty in the resulting SWE forecasts.

Using data from the 323 SNOTEL stations shown in Figure~\ref{fig:snotel_locs} spanning the 2001–2019 snow years, we show that an empirical implementation of the spatiotemporal whitening process yields competitive forecasting performance while conformal prediction provides reliable uncertainty estimates with empirical coverage near nominal levels.

%Thus, the primary contributions of this work are threefold. First, we extend GP-based spatial decorrelation to a spatiotemporal SWE forecasting setting. Second, we demonstrate that forecasting on decorrelated spatial residuals provides an effective framework for learning temporal SWE dynamics. Third, we incorporate conformal prediction to provide distribution-free uncertainty quantification with finite-sample coverage guarantees. We demonstrate the accuracy of this method in both standalone analysis and when compared to modern high-parameter approaches; we additionally offer distribution-free predictive intervals for water governance and statistical decision-making. 

The remainder of this paper is organized as follows. Section 2 reviews relevant background and related work. Section 3 presents the proposed methodology. Section 4 describes the data and experimental design. Section 5 presents reanalysis and uncertainty quantification results. Section 6 concludes with a discussion of findings, limitations, and future directions.

\section{Background and Related Work}

\subsection{Spatial Dependence in Statistical Learning}

A persistent challenge in applying machine learning to environmental data is accounting for spatial dependence among observations. GPs, which we define formally in Section~\ref{sec:spatial_whitening}, provide a flexible framework for representing spatial correlation, and allow for prediction at unobserved locations via a method known as Kriging \citet{cite-kriging}. However, the computational cost of working with GPs for large datasets can become prohibitive. Substantial effort has therefore been devoted to developing scalable approximations that retain the strengths of GP models while reducing computational burden. One such approach is the Nearest Neighbor Gaussian Process (NNGP) \citet{citation-datta2016}, which builds upon the conditional factorization proposed by \citet{cite-Vecchia}. In the NNGP, spatial dependence is represented through a sparse collection of local conditional relationships.

More recently, researchers have explored integrating machine learning into spatial statistical models. For example, \citet{cite-heaton} proposed a spatial whitening transformation framework that converts observations into approximately independent residuals prior to model training, allowing standard machine learning methods to operate on decorrelated data. To construct the whitening transformation, they use the NNGP approach \citet{citation-datta2016,cite-Katzfuss}. Alternatively, \citet{cite-zhandatta,zhan2026geospann} introduced the Neural Network Generalized Least Squares (NN-GLS) framework, which embeds neural-network mean functions within a spatial covariance model and estimates both jointly.

These developments highlight two complementary approaches for incorporating spatial dependence into machine learning: explicitly modeling spatial covariance within the learning algorithm or removing spatial dependence prior to model training. This work extends the latter strategy to a spatiotemporal SWE forecasting setting.

%Specifically, our proposed approach follows the whitening framework of \citet{cite-heaton} where unlike the classical Vecchia framework derived in \citet{cite-Vecchia}, where sparse nearest-neighbor constructions are introduced to approximate the GP likelihood, we use the nearest-neighbor graph solely to define a sparse spatial transformation. Therefore, the neighborhood construction is guided by the objective of preserving local spatial structure rather than by likelihood approximation. Additional details are provided in Section~\ref{sec:spatial_whitening}.
%More recently, researchers have explored integrating spatial statistical models with machine learning algorithms. For example, \citet{cite-heaton} proposed a spatial whitening transformation framework that converts observations into approximately independent residuals prior to model training, allowing standard machine learning methods to operate on decorrelated data. Alternatively, \citet{cite-zhandatta} introduced the Neural Network Generalized Least Squares (NN-GLS) framework, which embeds neural-network mean functions within a spatial covariance model and estimates both jointly. 

\subsection{Snow Water Equivalent Prediction}

Classical approaches to SWE prediction have often relied on spatial interpolation methods, such as Kriging, to estimate SWE at unobserved locations from nearby measurements \citep{cressie1993statistics, cite-carollcarollkriging}. These methods provide a principled framework for modeling spatial dependence at a fixed point in time, but do not address temporal snowpack dynamics.

%directly capture the temporal evolution of snow accumulation and melt over the course of a snow year. Thus, early spatiotemporal approaches extended these spatial models by incorporating temporal dynamics. For example, 
\citet{cite-huang1996} developed a spatiotemporal SWE modeling framework that combined Kriging with Kalman filtering to update predictions as new observations became available. This work demonstrated the value of jointly modeling spatial and temporal dependence in SWE prediction, although its linear structure limited flexibility in representing nonlinear relationships among SWE and its corresponding predictor variables.  

Recently, various DL methods have been applied to SWE forecasting. \citet{cite-SWEML} combines decision trees and multi-layer perceptrons (MLPs) to train 23 region-specific models across the Western United States. The resulting product---referred to as SWEML---provides daily snapshots of predicted SWE on a 1km-by-1km grid. \citet{cite-surveylidar} uses an MLP to estimate snow density, which can then be combined with aerial measurements of snow depth to construct a gridded SWE data product. We refer to \citet{ritchie2025benchmarking} for a comparison of these methods to other gridded SWE data products. 

Other works focus on point prediction of SWE using time series regression techniques. \citet{cite-Thapa} use a transformer \citet{cite-attention} to predict daily SWE at SNOTEL sites given only meteorological forcings and static attributes such as elevation. LSTMs, popularized for streamflow prediction by \citet{kratzert2018rainfall}, have also been applied in this setting \citet{meyal2020automated,duan2024using,song2024lstm,burns2026comparing}. However, none of these works jointly account for spatial correlation in conjunction with time series modeling. 

%Direct intercomparison between the aforementioned methods is challenging due to differing choices in training data, prediction resolution, and protocol (specifically: whether or not lagged SWE observations are provided to the model as an input). Consequently, we avoid doing so here, although future work could revisit this using the methodology of \citet{ritchie2025benchmarking}. 

\subsection{Uncertainty Quantification and Conformal Prediction}

Accurate point forecasts are important for SWE prediction, but many operational water-resource decisions also require reliable measures of predictive uncertainty. For example, reservoir management, drought planning, flood mitigation, and seasonal water-allocation decisions depend on both the expected amount of snowpack and the range of plausible future outcomes. Recent environmental forecasting studies have thus increasingly focused on interval prediction methods to better characterize predictive uncertainty and support risk-aware decision making \citep{iwakin2024conformal, xia2023interval}.

Quantifying uncertainty in DL models, however, remains challenging. Many existing approaches rely on distributional assumptions, Bayesian formulations, or ensemble methods, each of which may introduce additional computational complexity or require assumptions that are difficult to verify in practice \citep{gal2016dropout, lakshminarayanan2017simple}. Furthermore, uncertainty estimates produced by modern DL models are often poorly calibrated, particularly when applied to complex environmental systems \citep{guo2017calibration, angelopoulos2023conformal}.

Conformal prediction has emerged as a flexible framework for constructing prediction intervals with finite-sample coverage guarantees under relatively weak assumptions \citep{vovk2005algorithmic, shafer2008tutorial}. Conformal methods use a held-out calibration dataset to estimate the distribution of prediction errors and construct prediction intervals accordingly rather than requiring a fully specified probabilistic model,  \citep{lei2018distributionfree, angelopoulos2023conformal}. This distribution-free and model-agnostic property has led to growing interest in conformal prediction across a wide range of machine learning applications \citep{angelopoulos2023conformal, kaiser2025tutorial}.

\section{Proposed Methodology}
\label{sec:proposed_methodology}

We build upon the spatial whitening transformation of \citet{cite-heaton}, which we discuss in Section~\ref{sec:spatial_whitening}. Our proposed extension of this methodology to spatiotemporal data is presented in Section~\ref{sec:spatiotemporal}. We discuss how we incorporate conformal prediction into our framework in Section~\ref{sec:methodology_conformal_prediction}. Along the way, we discuss various refinements we have made to the original spatial transformation codebase as provided in \citet{cite-heaton}. While we ground our methodological contributions using the running example of SWE prediction, our techniques easily transfer to other types of episodic spatiotemporal data.

\subsection{Notation}
\label{sec:Notation}
Throughout we consider a dataset
\begin{equation*}
\{y_{i,a,t} = y(\bm{s}_i,a,t) : i = 1, \dots, N,\; t = 1, \dots, T, a \in \{a_{\min},\ldots, a_{\max}\} \}  
\end{equation*}
of a response variable observed at spatial locations $\bm{s}_i \in \mathbb{R}^2$ during snow year $a$ and at day $t \in \{1,\ldots T\}$ within this snow year. Note that $\{a_{\min},a_{\max}\}$ are the first and last available snow years within the dataset, respectively. In our setting $a_{\min} = 2001$ and $a_{\max} = 2019$. We also consider features or covariates observed at the same locations, snow years, and days: $\{x_{i,a,t} = x(\bm{s}_i,a,t)\}$. For any $\bm{\Omega} \in \mathbb{R}^{N\times N}$ and any index sets $A,B \subseteq \{1,\ldots,N\}$, we denote by $\bm{\Omega}(A,B)$ the submatrix of $\bm{\Omega}$ formed by the rows indexed by $A$ and the columns indexed by $B$. Similarly, $\bm{\Omega}_{A}$ denotes the submatrix of $\bm{\Omega}$ consisting of the rows indexed by $A$ and all columns.  When either $A$ or $B$ is a singleton, we write $i$ in place of the singleton set $\{i\}$. 

\subsection{The Spatial Whitening Transformation}
\label{sec:spatial_whitening}

\subsubsection{The Geostatistical Model}
For now, let us ignore temporal dependence by keeping $a$ and $t$ fixed. Consider the random vector $\bm{y}=[y(\bm{s}_1),\ldots,y(\bm{s}_N)]^\top\in\mathbb{R}^N$
observed at spatial locations $\{\bm{s}_1,\ldots,\bm{s}_N\}$. Let $\bm{x}(\bm{s})\in\mathbb{R}^F$ denote the feature vector at location $\bm{s}$, and define the corresponding design matrix
\[
\bm{X}
=
\begin{bmatrix}
\bm{x}(\bm{s}_1)^\top \\
\vdots \\
\bm{x}(\bm{s}_N)^\top
\end{bmatrix}
\in \mathbb{R}^{N \times F}.
\]
As is standard in spatial statistics \citep{cressie1993statistics, banerjee2014hierarchical}, we model the response process $y(\bm{s})$ observed at locations $\{\bm{s}_1,\ldots,\bm{s}_N\}$ using the additive geostatistical model
\[
y(\bm{s}_{i}) = \bm{x}(\bm{s}_{i})^\top\bm{\beta}
+g(\bm{s}_{i})+\varepsilon_{i}.
\]
Here $\bm{\beta}\in\mathbb{R}^F$ denotes the vector of unknown regression coefficients, $g(\bm{s})$ is a stationary, mean-zero GP with marginal variance $\sigma^2$, and the $\varepsilon_i$ represent independent Gaussian measurement errors with mean zero and variance $\tau^2$. %$\tau^2$ is commonly referred to as the nugget effect, accounting for measurement error and microscale spatial variation that is not captured by the latent spatial process $g(\bm{s})$.
Therefore, 
\begin{equation}
\bm{y}\sim \mathcal{N}(\bm{X}\bm{\beta},\Sigma),\ \mathrm{where}\  \Sigma = \sigma^{2} R 
\label{eq:Sigma_defn}
\end{equation}
In \eqref{eq:Sigma_defn}, $K \in \mathbb{R}^{N \times N}$ is the spatial correlation matrix with entries $K_{ij}=\kappa\left(||\bm{s}_i-\bm{s}_j||;\bm{\alpha}\right)$,
where $\kappa(\cdot;\bm{\alpha})$ is a valid correlation function parameterized by $\bm{\alpha}$. We parameterize the covariance using a nugget proportion $\eta$ such that $R=(1-\eta)K+\eta I$; equivalently, under the traditional decomposition into spatial variance and nugget variance, $\tau^2/\sigma^2=\eta/(1-\eta)$. Throughout this paper, we adopt the Mat\'ern correlation function,
\[
\kappa(r)=
\frac{1}{\Gamma(\nu)\,2^{\nu-1}}
\left(\frac{r\sqrt{2\nu}}{\rho}\right)^{\nu}
\Psi_{\nu}\left(\frac{r\sqrt{2\nu}}{\rho}\right),
\quad r \ge 0,
\]
where $\Psi_{\nu}(\cdot)$ denotes the modified Bessel function of the second kind of order $\nu$. The parameter $\rho>0$ controls the spatial correlation range, while $\nu>0$ governs the smoothness of $g(\cdot)$. The Mat\'ern family provides a flexible class of covariance models that can represent a wide range of spatial dependence structures and is widely used in spatial statistics \citet{stein1999interpolation}.
%where $R \in\mathbb{R}^{N\times N}$ is the %spatial correlation matrix, such that $R_{ij}
%    =
%    \kappa\!\left(\|s_i-%s_j\|;\bm{\alpha}\right)$.
%Here, $\kappa(\cdot;\bm{\alpha})$ is a %correlation function with parameters
%$\bm{\bm{\alpha}}$. Throughout we shall take %$\kappa (\cdot)$ to be the Mat\'ern kernel:
%\begin{equation*}
%    \kappa(r) = \frac{1}{\Gamma(\nu)\,2^{\nu-%1}}\left(\frac{r\sqrt{2\nu}}%{\rho}\right)^{\nu}\Psi_%{\nu}\!\left(\frac{r\sqrt{2\nu}}{\rho}\right) \quad \text{ for } r \in \mathbb{R},
%    \label{matern}
%\end{equation*}
%where, $\rho>0$ controls the spatial correlation range, $\nu>0$ controls smoothness, and $\Psi_{\nu}$ denotes a modified second-order Bessel function. see \citet{stein1999interpolation} for a discussion on this choice. %The parameter $\sigma^2$, which appears through $\Sigma=\sigma^2R$, is identical to the variable $\sigma^2$ defined in Equation~\ref{eq:Sigma_defn}.\\

\subsubsection{The Vecchia Approximation}
 Let $p(\bm{y})$ denote the probability density function associated to $\bm{y}$. This density can be factored as
$$
    p(\bm{y}) = p(y(\bm{s}_1)) \prod_{i=2}^n p(y(\bm{s}_i) | y(\bm{s}_{i-1}), \dots, y(\bm{s}_1)),   
$$
where each term in the product is a Gaussian density function that is conditioned on the preceding variables. Vecchia \citet{cite-Vecchia} introduced a computationally efficient approximation to $p(\bm{y})$ by simply restricting the conditioning to a nearest-neighbor set. The classical Vecchia approximation is constructed by first imposing an ordering $\prec$ on the spatial locations and then restricting each conditioning set to previously ordered locations \citet{citation-datta2016, cite-Katzfuss}. In this
construction, the approximation
\begin{equation}
    p(\bm{y})
    \approx
    \breve{p} (\bm{y})
    =
    p(y(\bm{s}_1))
    \prod_{i=2}^{N}
    p\left(y(\bm{s}_i)\mid y(\bm{s}_j),\, j\in\mathcal C_i\right),
    \label{eq:Vecchia_approx}
\end{equation}
defines a valid probability density, where each conditioning set $\mathcal C_i$ (with $\mathcal{C}_{1} = \varnothing$) consists only of predecessors under the chosen ordering.

\subsubsection{Choosing the Conditioning Sets}
Among the many possible orderings, the maximin ordering has been shown to provide improved approximation accuracy for large spatial datasets and is generally recommended \citet{cite-Katzfuss, guinness2018}. In this work, however, we construct the neighborhood structure using the $M \sim 30$ nearest spatial locations {\em without} imposing a global ordering. This choice is motivated by two dominant characteristics of the SNOTEL dataset.

First, this data set is of moderate size, containing only $323$ monitoring sites. Enforcing an ordering would substantially restrict the available conditioning sets. In particular, locations appearing early in the ordering necessarily have fewer than $M$ predecessors available, while for many subsequent locations the nearest predecessor locations under a maximin ordering may still be relatively far from the target location because the ordering is designed to spread points throughout the spatial domain. Second, there is pronounced local spatial continuity in snowpack observations due to the influence of local topographic and meteorological conditions such as elevation, slope, aspect, and prevailing climate \citet{carroll1997spatial,fassnacht2003snow,dozier2016estimating}. Since the spatial transformation proposed here is used solely as a computational device rather than as a likelihood approximation for a latent GP, our primary objective is to preserve these local spatial relationships. %Therefore, for the moderately sized SNOTEL network considered here, we construct each neighborhood directly from its nearest spatial locations instead of restricting conditioning sets to predecessor locations under a global ordering. %Although ordered Vecchia approximations possess attractive theoretical properties and maximin ordering has been shown to improve approximation accuracy for large spatial datasets \citet{cite-Katzfuss,guinness2018}, preserving geographically local neighborhoods is more appropriate for the spatial transformation considered in this work. Nevertheless, 

We note that we can straightforwardly modify our approach to incorporate a global ordering by updating the nearest-neighbor search routine so that, for each location $\bm{s}_i$, the code filters the candidate neighbors to those satisfying $\bm{s}_j \prec \bm{s}_i$ before selecting the $M$ nearest locations.

\subsubsection{Whitening the Data: Ideal Case}
Suppose that $\bm{y}$ is sampled \textit{exactly} from the distribution on the right-hand side of \eqref{eq:Vecchia_approx}. Then as pointed out in \citet{cite-heaton}, $y(\bm{s}_i) \sim \mathcal{N}(\mu_i,\sigma^2\nu_i)$ where
\begin{align*}
    \mu_i &= \left\{\begin{array}{cc} \bm{x}(\bm{s}_i)^{\top}\bm{\beta} & \text{ if } i = 1 \\ \bm{x}(\bm{s}_i)^{\top}\bm{\beta} + R(i,\mathcal{C}_i)R^{-1}(\mathcal{C}_i,\mathcal{C}_i)(y_{\mathcal{C}_i} - X_{C_i}\bm{\beta}) & \text{ if } i > 1\end{array}\right. \\
    \nu_i &= \left\{\begin{array}{cc}1 & \text{ if } i=1 \\ 1 - R(i,\mathcal{C}_i)R^{-1}(\mathcal{C}_i,\mathcal{C}_i)R(\mathcal{C}_i,i) & \text{ if } i > 1 \end{array}\right..
\end{align*}
Since $\Sigma=\sigma^2R$, the corresponding covariance submatrix satisfies
$\Sigma(A,B)=\sigma^2R(A,B)$. For $i > 1$ we define the \textit{Kriging weights} \citet{cressie1993statistics} by
\begin{equation}
\bm{w}_i^{\top}
= R(i,\mathcal{C}_i)
R(\mathcal{C}_i,\mathcal{C}_i)^{-1}
\in\mathbb{R}^{m_i},
\quad \text{ where } m_i = |\mathcal{C}_i|.
\label{eq:define_kriging_weights}
\end{equation}
This is the vector of coefficients for the best linear predictor of  $y(\bm{s}_i)$ given $y(\bm{s}_j)$ for $j \in \mathcal{C}_i$. Since $\mathcal{C}_1=\varnothing$, no Kriging-weight vector is defined for the first ordered location. %For $i>1$, the dimension of $\bm{w}_i$ is $m_i=|\mathcal{C}_i|$, corresponding to the number of locations in the conditioning set.
%
%Let us also define the {\em Kriging weights} \citet{cressie1993statistics,bandyopadhyay_spatial}
%\begin{equation}
%    \bm{w}_i^{\top}
%    =
%    R(i,\mathcal{C}_i)
%    R(\mathcal{C}_i,\mathcal{C}_i)^{-1}
%    \in\mathbb{R}^{m_i},
%    \quad i>1.
%    \label{eq:define_kriging_weights}
%\end{equation}
%Because $\mathcal C_1=\varnothing$, no Kriging-weight vector is required for the first ordered location. For $i>1$, the dimension of $\bm w_i$ matches the number $m_i$ of locations in $\mathcal C_i$.
%
% The same weights result if we use the full covariance matrix $\Sigma=\sigma^2R$, because the common factor $\sigma^2$ cancels:
% \[
%     \Sigma(i,\mathcal C_i)
%     \Sigma(\mathcal C_i,\mathcal C_i)^{-1}
%     =
%     R(i,\mathcal C_i)
%     R(\mathcal C_i,\mathcal C_i)^{-1}.
% \]
%
For $i>1$,  we can thus rearrange the formula for $\mu_i$ and $\nu_i$ as
\begin{equation}
\begin{split}
    \mu_i
    &=
    \left(\bm{x}(\bm{s}_i)^{\top} -\bm{w}_i^{\top}X_{\mathcal{C}_i}\right)\bm{\beta}
    + \bm{w}_i^{\top}y_{\mathcal{C}_i} \\
    \nu_i &= 1 - \bm{w}_i^{\top}R(\mathcal{C}_i,i).
\end{split}
    \label{eq:kriging_weights_mu_nu}
\end{equation}
and define,
\begin{equation}
\begin{split}
    \widetilde{\bm{x}}(\bm{s}_i) &= \nu_i^{-1/2}\left(\bm{x}(\bm{s}_i) - X_{\mathcal{C}_i}^{\top}\bm{w}_i\right) \\
    \widetilde{y}(\bm{s}_i) &= \nu_i^{-1/2}\left(y(\bm{s}_i) - \bm{w}_i^{\top}y_{\mathcal{C}_i}\right)
\end{split}
    \label{eq:kriging_weights_x_Y}
\end{equation}
with $\widetilde{\bm{x}}(\bm{s}_1) = \bm{x}(\bm{s}_1)$ and $\widetilde{y}(\bm{s}_1) = y(\bm{s}_1)$. Following \citet{cite-heaton}, these transformed quantities are referred to as the {\em spatially whitened covariates} and {\em spatially whitened responses}, respectively. As observed in \citet{cite-heaton}, $\widetilde{\bm{y}} \sim \mathcal{N}(\widetilde{\bm{X}}\bm{\beta},\sigma^2I)$, 
i.e., the whitened response is spatially uncorrelated, and has mean determined by the spatially whitened covariates while retaining the marginal variance. Consequently,  the ordinary least squares (OLS) estimator 
\begin{equation}
    \widehat{\bm{\beta}} = \operatorname*{argmin}_{\bm{\alpha}\in\mathbb{R}^F}\|\widetilde{\bm{y}} - \widetilde{\bm{X}}\bm{\alpha}\|^2 = \operatorname*{argmin}_{\bm{\alpha}\in\mathbb{R}^F}\sum_{i=1}^N\left(\widetilde{y}(\bm{s}_i) - \widetilde{\bm{x}}(\bm{s}_i)^{\top}\bm{\alpha} \right)^2
    \label{eq:OLS}
\end{equation}
is optimal by the Gauss-Markov theorem.

\subsubsection{Whitening the Data: Real-World Case}
If $\bm{y}$ is not sampled exactly from the Vecchia-approximated distribution (i.e., the right-hand side of \eqref{eq:Vecchia_approx}; see Appendix~\ref{app:Gaussianity} for a discussion of Gaussianity in the present application) {\em or} the dependence of the mean on the covariates is non-linear (i.e. not of the form $\bm{x}(\bm{s})^{\top}\bm{\beta}$ in the untransformed coordinate) then the use of OLS is no longer strictly theoretically justified. Nevertheless, \citet{cite-heaton} argue that spatial whitening ``mitigates the effects of spatial correlation during model training'' and advocate for training DL models using the spatially whitened observations,
\begin{equation}
    \widehat{f} = \operatorname*{argmin}_{f \in \mathcal{F}}\sum_{i=1}^N\left(\widetilde{y}(\bm{s}_i) - f(\widetilde{\bm{x}}(\bm{s}_i))\right)^2,
    \label{eq:OLS_training_problem}
\end{equation}
where $\mathcal{F}$ is a suitable hypothesis class (e.g., deep neural networks of a specified architecture). They present convincing numerical evidence that this approach does indeed boost model performance. Notably, \eqref{eq:OLS_training_problem} allows for standard DL tricks such as mini-batching, as each summand within the loss function is independent. 

After training, predictions at a new location $\bm{s}$ are made using $\widehat{f}$ as follows. First, a nearest neighbors set $\mathcal{C}_{\bm{s}}\subset\{1,\ldots, N\}$ is computed as above. Next, the Kriging weights are computed as 
$\bm{w}_{\bm{s}}^{\top} = \Sigma(\bm{s},\mathcal{C}_{\bm{s}})\Sigma^{-1}(\mathcal{C}_{\bm{s}},\mathcal{C}_{\bm{s}})
$
following which $\nu_{\bm{s}}$ and $\widetilde{\bm{x}}(\bm{s})$ are computed using formulas \eqref{eq:kriging_weights_mu_nu} and \eqref{eq:kriging_weights_x_Y}, suitably adapted. The model prediction $\widehat{f}(\widetilde{\bm{x}}(\bm{s}))$ is then computed and recorrelated:
\begin{equation}
  \widehat{y}(\bm{s}) = \nu_{\bm{s}}^{1/2}\widehat{f}(\widetilde{\bm{x}}(\bm{s})) + \bm{w}_{\bm{s}}^{\top}\bm{y}_{{\mathcal{C}_{\bm{s}}}}.
  \label{eq:recalibration}
\end{equation}

\subsection{Decorrelation for Spatiotemporal Data}
\label{sec:spatiotemporal}

We now extend the spatial whitening transformation to the spatiotemporal setting.  %Next, we extend the spatial whitening transformation to spatiotemporal data by leveraging the fact that the spatial locations $\bm{s}_i$ are not changing with time which means the Kriging weights defined in \eqref{eq:define_kriging_weights} and used in \eqref{eq:kriging_weights_mu_nu} need only be computed once and stored. 
Defining  $\bm{y}_{a,t} = [y(\bm{s}_1,a,t),\ldots, y(\bm{s}_N,a,t)]^{\top}$ and $\bm{X}_{a,t} = \left[\bm{x}(\bm{s}_1,a,t) \cdots \bm{x}(\bm{s}_N,a,t)\right]^{\top}$ we compute the spatially whitened spatiotemporal data as 
\begin{equation}
\begin{split}
    \widetilde{\bm{x}}(\bm{s}_i,a,t) &= \nu_i^{-1/2}\left(\bm{x}(\bm{s}_i,a,t) - X_{a,t,\mathcal{C}_i}^{\top}\bm{w}_i\right) \\
    \widetilde{y}(\bm{s}_i,a,t) &= \nu_i^{-1/2}\left(y(\bm{s}_i,a,t) - \bm{w}_i^{\top}y_{a,t,\mathcal{C}_i}\right) 
\end{split}
\label{eq:kriging_weights_x_Y_temporal}
\end{equation}
for $i=2,\ldots,N$. We assume that the spatial locations $\{\bm{s}_1,\ldots,\bm{s}_N\}$ remain fixed over time. Therefore, the Kriging weights ($\bm{w}_i$, see \eqref{eq:define_kriging_weights}) and the conditional variances ($\nu_i$, see \eqref{eq:kriging_weights_mu_nu}) are independent of both $a$ and $t$ and need only be computed once. The resulting weights can then be stored and reused at every time point, making the whitening transformation computationally efficient for long spatiotemporal sequences. To improve scalability further, we use k-d trees for efficient nearest-neighbor searches and the Cholesky decomposition to compute the Kriging weights in \eqref{eq:define_kriging_weights}. Additional implementation details are provided in Appendix~\ref{app:cholesky}.
%Note the $\nu_i$ are also time independent, and so also only need to be computed once (see \eqref{eq:kriging_weights_mu_nu}). We incorporate several additional techniques for scalability, notably the use of k-d trees for nearest neighbors computations and the Cholesky decomposition for computing the $\bm{w}_i$ using \eqref{eq:define_kriging_weights}. See Appendix~\ref{app:cholesky} for further details.Accordingly, a prediction on the original scale is obtained from

\subsection{Training for Spatiotemporal Data}
\label{sec:spatiotemporal_training}
Recent benchmarking supports the use of LSTMs for hydrologic prediction. Specifically, \citet{cite-LiuRNNTransformers} found that LSTMs generally outperformed attention-based models in regression and short-range forecasting tasks. This is partially due to LSTM gating mechanisms (discussed in more detail in Appendix~\ref{app:lstm}), which help preserve both short- and long-term temporal dependence. While transformer-based models showed greater advantages mainly for more complex longer-horizon autoregressive prediction, for hydrologic prediction, the best transformer architecture only slightly outperformed the LSTM, indicating that LSTMs remain a competitive choice for the present task when balancing parameter counts and training time and complexity. 

For the forecasting model $f$ we use an LSTM in sequence-to-sequence mode; concretely, this means that 
\begin{equation}
    f:\mathbb{R}^{T\times F}\rightarrow \mathbb{R}^{T},
    \label{eq:forecastingmodel}
\end{equation}
where $T$ is the number of days in a snow-year and $F$ is the number of features. For each station $\bm{s}_i$ and snow year $a$, let
\[
\widetilde X_{i,a}
=
\begin{bmatrix}
\widetilde x(\bm{s}_i,a,1)^{\top} \\
\widetilde x(\bm{s}_i,a,2)^{\top} \\
\vdots \\
\widetilde x(\bm{s}_i,a,T)^{\top}
\end{bmatrix}
\in \mathbb{R}^{T\times F}
\]
denote the sequence of whitened features. The model returns a sequence of decorrelated SWE predictions
\[
\widehat{\widetilde{\bm y}}_{i,a}
=
f(\widetilde X_{i,a})
\in \mathbb{R}^{T},
\]
where the $t$-th component $\widehat{\widetilde y}_{i,a,t}$ corresponds to the predicted decorrelated SWE value at station $\bm{s}_i$ on day $t$ of snow year $a$. Let $\mathcal{A}_{\mathrm{train}}$ denote the set of snow years included in the training data. We train by minimizing mean squared error (MSE) over all locations, snow years within the training set, and days:
\begin{equation}
    \widehat f=\operatorname*{argmin}_{f\in\mathcal{F}}
    \frac{1}{NT|\mathcal{A}_{\mathrm{train}}|}
    \sum_{a\in \mathcal{A}_{\mathrm{train}}}
    \sum_{i=1}^{N}
    \left\|\widehat{\widetilde{\bm y}}_{i,a}-
        \widetilde{\bm y}_{i,a}
    \right\|^2.
    \label{eq:LSTM_MSE}
\end{equation}
We construct minibatches by sampling a particular $a$, and a subset of the $N$ locations, but we do not split up the predictions within a particular snow year. After training, model predictions are recorrelated using \eqref{eq:recalibration}. 

\subsection{Model Evaluation}
SWE prediction is often quantified using Nash-Sutcliffe Efficiency (NSE) \citet{cite-NSE}:
\begin{equation}
    \text{NSE}_{i,a} = 1 - \frac{\sum_{t=1}^{T} \big(y_{i,a,t} - \widehat{y}_{i,a,t}\big)^2}{\sum_{t=1}^{T} \big(y_{i,a,t} - \overline{y}_{i,a,t}\big)^2},
    \label{eq:nse}
\end{equation}
where $\overline{y}_{i,a}$ denotes the snow-year mean at station $i$. NSE values range from $(-\infty, 1]$, with values closer to one implying strong model performance and $\text{NSE} \le 0$ implying the model's predictions are worse than using the mean as a predictive measure. %In other words, NSE is a measure of how well a model's predictions compare to the daily mean for a given day $t \in \{1, \dots, T\}$. This metric forms the foundation for model analysis, with values greater than $0.5$ indicating strong model performance in practice.

We evaluate forecasting performance on the held-out test dataset using root mean squared error (RMSE) and NSE. We compare against a climatological baseline that predicts $\hat{y}_{i,a,t}^{\text{clim}}$ to be the historical mean for that station and day:
\begin{equation*}
    \hat{y}_{i,a,t}^{\text{clim}} = \frac{1}{|\mathcal{A}_{\mathrm{train}}|} \sum_{b\in \mathcal{A}_{\mathrm{train}}}y_{i,b,t}.
\end{equation*}

\subsection{Conformal Prediction}
\label{sec:methodology_conformal_prediction}

\subsubsection{Implementation}
We use split conformal prediction~\citep{lei2018distributionfree, Papadopoulos2001} to quantify uncertainty in the point forecasts produced by the fitted LSTM model. To do so, we partition the available snow years into disjoint training, validation, calibration, and testing sets. We use the training set to estimate the model parameters, the validation set to select the best training epoch, the calibration set to construct the conformal prediction intervals, and the testing set to evaluate the resulting point forecasts and empirical interval coverage.

After applying the spatial back-transformation, for each station $s_i$ and snow year $a$ the forecasting model produces
\[
\widehat{\bm{y}}_{i}^a =
\left( \widehat{y}_{i,a,1},
\ldots,
\widehat{y}_{i,a,T}
\right)^{\top}
\in \mathbb{R}^{T},
\label{eq:cp_prediction_vector}
\]
where $\widehat{y}_{i,a,t}$ denotes the predicted SWE on the original scale at station $s_i$ on snow day $t$ of snow year $a$. Let
$\mathcal{A}_{\mathrm{cal}}
\subseteq
\{{a_{\min},\ldots,a_{\max}}\}$ denote the set of snow years in the calibration data. For each $a\in\mathcal{A}_{\mathrm{cal}}$, station $i\in\{1,\ldots,N\}$, and snow day $t\in \{1,\ldots,T\}$, we define the nonconformity score
\[
r_{i,a,t}
=
\left|
y_{i,a,t}
-
\widehat{y}_{i,a,t}
\right|,
\label{eq:cp_nonconformity}
\]
which measures the absolute forecasting error associated with the given calibration observation.

We estimate conformal quantiles separately for each snow day. This accounts for the fact that SWE variability and forecasting difficulty differ substantially between accumulation, peak, and melt periods. For a fixed snow day $t\in\{1,\ldots,T\}$, define
\[
\mathcal{R}_t
=
\left\{
r_{i,a,t}
:
i\in\{1,\ldots,N\},
a\in\mathcal{A}_{\mathrm{cal}}
\right\}.
\label{eq:cp_residual_set}
\]
$\mathcal{R}_t$ contains the calibration residuals associated with snow day $t$, pooled across all stations and calibration snow years. Importantly, pooling across stations increases the number of calibration residuals available for each snow day while still allowing interval widths to adapt to seasonal changes in predictive uncertainty. 

More specifically, let $m_t =\left|\mathcal{R}_t\right|$ denote the number of available calibration residuals for snow day $t$, and let $r_{t,(1)}\leq r_{t,(2)}\leq\cdots\leq r_{t,(m_t)}$ denote the corresponding ordered residuals. For a prescribed nominal coverage level $(1-\alpha)\in(0,1)$, define
\begin{equation}
k_t
=
\left\lceil
(m_t+1)(1-\alpha)
\right\rceil.
\label{eq:cp_rank}
\end{equation}
We then define the snow-day-specific conformal quantile as $q_{1-\alpha}(t) = r_{t,(k_t)}$. For a test observation at station $s_i$, snow year $a$, and snow day $t$, we construct the conformal prediction interval
\begin{equation}
    C_{\alpha}(i,a,t)
=
\left[
\widehat{y}_{i,a,t}
-
q_{1-\alpha}(t),
\widehat{y}_{i,a,t}
+
q_{1-\alpha}(t)
\right].
\label{eq:cp_interval}
\end{equation}
The conformal half-width therefore varies throughout the snow year but remains common across stations for a fixed snow day, allowing the interval width to reflect seasonal changes in forecasting uncertainty.

\subsubsection{Exchangeability}
A key component of conformal prediction is that observations $y_{i,a,t}$ within the calibration and test sets must be \textit{exchangeable} (and thus the resulting nonconformity scores); in other words, their joint distribution remains unchanged regardless of permutation:
\[
r_{1,a,t}, \dots, r_{N,a,t} \overset{d}{=} r_{\phi(1),a,t}, \dots, r_{\phi(N),a,t}
\]
for all permutations $\phi$ and all snow years $a$ in the calibration and test sets. If, for a fixed snow day $t$, the nonconformity scores for all stations $\bm{s}_i$ and snow years $a$ are exchangeable, then the split-conformal construction provides the marginal coverage guarantee
\begin{equation}
\mathbb{P}
\left(
y_{i,a,t}
\in
C_{\alpha}(i,a,t)
\right)
\geq
1-\alpha.
\label{eq:cp_coverage}
\end{equation}  
We perform conformal prediction after transforming predictions back to the original SWE scale; therefore, the exchangeability assumption applies to the resulting prediction errors. However, the spatiotemporal dependence among stations, snow years, and successive timesteps prevents us from establishing the theoretical guarantees that underly the exchangeability conditions required for the standard finite-sample coverage guarantee. Accordingly, the coverage result in Equation~\eqref{eq:cp_coverage} relies on an assumption that may hold only approximately for the present spatiotemporal data. %Accordingly, we view the proposed conformal prediction procedure as an exploratory implementation intended to assess the practical behavior of split conformal prediction within a hydrological forecasting framework.

\subsubsection{Evaluation}
\label{sec:PICP}
We evaluate the efficacy of the prediction intervals defined in Equation~\eqref{eq:cp_interval} using the prediction interval coverage probability (PICP).  For each test prediction $\widehat{y}_{i,a,t}$, we determine whether the corresponding observation $y_{i,a,t}$ falls within the interval $C_{\alpha}(i,a,t)$. We then compute the PICP as the proportion of test observations contained within their corresponding prediction intervals.

\section{Data and Experimental Design}
We considered candidate data from daily SNOTEL observations  for 822 stations across 19 snow years (2001-2019) \citet{cite-SNOTELdata}. To avoid missingness, we removed stations with missing values in any variable or year. The resulting 323 locations comprised our primary data set. 
\subsection{Experimental Features}
\label{subsec:features}

The set of static features and meteorological forcings used for model training are as follows:
\begin{itemize}
    \item \underline{In Situ SWE Measurements:} Ground truth observations from 323 SNOTEL stations across the Western United States, as shown in Figure~\ref{fig:snotel_locs}, provide SWE values used to train and evaluate the model.
    \item \underline{Static Features:} Elevation, latitude, longitude, slope \citet{cite-SNOTELdata}, land cover \citet{cite-yang}
    \begin{itemize}
        \item The landcover grid was at a 30m resolution with the dominant landcover type from the grid containing each location extracted for data analysis. 
        \item Elevation and slope were derived in ArcGIS from a 30m Digital Elevation Model (DEM), following \citet{cite-Thapa}.
    \end{itemize}
    \item \underline{Meteorological Forcings:} Maximum Daily Temperature, Minimum Daily Temperature, Average Observed Daily Temperature, Daily Precipitation \citet{cite-SNOTELdata}
    \item \underline{Satellite Measurements:} Passive Microwave Brightness Temperature (19/37 GHz) 
    \begin{itemize}
        \item Passive microwave observations are sensitive to snowpack properties such as depth, grain size, and liquid water content.
        \item The 19 GHz and 37 GHz channels respond differently to scattering within the snowpack, and their difference has been widely used to estimate snow water equivalent and snow depth in remote sensing studies (e.g., \citet{cite-brodzik}).
    \end{itemize}
    \item \underline{Day-of-Snow-Year,} a single scalar variable which transforms the date of the calendar year to the day of the snow year (October 1 = 1, October 2 = 2, $\dots$, June 30 = 273). 
\end{itemize}

This project's snow year timeframe for which to make predictions spans October 1 to June 30, a period of 273 days. This yielded $323 \times 19 \times 273 = 1,675,401$ total [location, year, day] data points. We split the data into training, validation, and testing subsets corresponding to distinct snow years. Specifically, we use 2001--2012 for training, 2013--2015 for validation, and 2016--2019 for testing. When performing conformal prediction, we use 2014--2015 as the calibration dataset and adjust the training and validation sets to 2001--2011 and 2012--2013, respectively, while keeping the test years unchanged.  

%With regards to data availability, meteorological forcings and static features are broadly available from both in-situ measurements and gridded reanalysis products, enabling both spatial interpolation and future scenario modeling. Passive microwave measurements are globally available via satellite platforms, making them particularly valuable for extending SWE estimation to ungauged or data-sparse regions.

% \subsection{Model Setup}
% We split the 19 years into three sets:
% \begin{itemize}
%     \item Training Data: 2001-2012
%     \item Validation Data: 2013-2015
%     \item Testing Data: 2016-2019
% \end{itemize}

% The dataset is partitioned into s . The training dataset is used for model estimation, the validation dataset is used for model selection, and the test dataset is reserved for final out-of-sample evaluation. This design preserves the temporal structure of the data while ensuring that model performance is evaluated on snow years that were not observed during training. We also note that when implementing conformal prediction, we further separate the data into four different sets, including a calibration set.

\subsection{Data Preparation}

We merge all static and dynamic covariates into a common spatiotemporal dataset and partition the resulting observations into individual snow years. Next, we apply the spatial whitening transformation described in Section~\ref{sec:spatial_whitening} to SWE observations and dynamic meteorological forcings. We cache the Kriging weights $\bm{w}_i$ for later use. %This transformation produces spatially decorrelated variables together with the quantities required to reconstruct predictions on the original spatial scale.

After spatial whitening, we apply z-score normalization to the  predictor variables and SWE observations by subtracting the training set mean and dividing by the training set standard deviation. We then organize the resulting decorrelated and normalized sequences into station-year samples that serve as inputs to the forecasting model. %Consequently, the LSTM learns temporal dynamics from spatially whitened observations while preserving the information required to reconstruct forecasts in the original coordinate system.

%We implement spatial whitening using a nearest-neighbor approximation to the GP covariance structure. For each station, we identify the $M$ nearest neighboring locations and use these neighbors to construct the whitening transformation described in Section~\ref{sec:spatial_whitening}. To improve computational efficiency, we use KD-trees for nearest-neighbor search and cache the resulting whitening quantities so that the same station-level transformation can be reused across timesteps. We apply this transformation to every timestep, yielding a decorrelated spatiotemporal dataset that forms the basis for model training, validation, and testing.

\subsection{Custom Configuration Arguments for Model Training}

We leverage the AdamW optimizer \citet{loshchilov2017decoupled} for model training with default values for the momentum parameters ($\beta_1 = 0.9,\beta_2 = 0.999$). We used a learning rate scheduler---{\tt ReduceLROnPlateau} in Pytorch---with an initial learning rate of $0.001$. Input variables are normalized prior to training, and model architecture is controlled through standard LSTM hyperparameters set by the user, including the hidden-state dimension, number of recurrent layers, and batch size. For the final LSTM architecture, we used a hidden-state dimension of $256$, four recurrent layers, and a batch size of $512$, with  mean squared error (MSE) as the loss function. We evaluate model performance using root mean squared error (RMSE) and NSE.

In addition to the forecasting model hyperparameters, our framework includes several parameters governing the spatial whitening transformation: the Matérn covariance parameters, the nugget proportion, and the number of nearest neighbors used in the Vecchia approximation. We selected these through manual experimentation; specifically, we cycled through several combinations and compared the resulting predictive performance until we identified an appropriate combination. For the final model, we fixed a Matérn smoothness parameter of $\nu=1.0$, range parameter of $\rho=2.0$, nugget proportion of $\eta = 0.1$, and $M=10$ nearest neighbors in the Vecchia approximation.

\section{Results}
\label{sec:results}
\subsection{Point Forecast Performance}

Our spatiotemporal model outperformed the climatological baseline on both evaluation metrics. Furthermore, the improvement over the baseline was broadly consistent across stations. The spatiotemporal whitening framework achieved NSE values greater than 0.75 at 240 of the 323 SNOTEL stations, compared with 121 stations for the climatological baseline. Conversely, only three stations produced negative NSE values, whereas the climatological baseline yielded negative NSE values at eight stations. %Together, these results indicate that the proposed spatiotemporal whitening framework provides accurate and consistent point forecasts across a diverse range of snow climates and physiographic settings.

\begin{table}[h]
\centering
\caption{Comparison of overall forecasting performance and station-level NSE distributions between the spatiotemporal model and the climatological baseline.}
\vspace{3mm}
\label{baseline-results-comparison}
\begin{tabular}{lcc}
\hline
\textbf{NSE Range} & \textbf{This Study} & \textbf{Climatology Baseline} \\
\hline
$\text{NSE} \le 0$               & 3   & 8   \\
$0 < \text{NSE} \le 0.3$         & 4   & 11  \\
$0.3 < \text{NSE} \le 0.5$       & 9  & 54  \\
$0.5 < \text{NSE} \le 0.75$      & 67  & 129 \\
$0.75 < \text{NSE} \le 1.0$      & 240 & 121 \\
\hline
\multicolumn{1}{r}{\textbf{Overall NSE}} & \textbf{0.877} & \textbf{0.777} \\
\hline
\multicolumn{1}{r}{\textbf{Overall RMSE}} & \textbf{3.84 in.} & \textbf{5.2 in.} \\
\hline
\end{tabular}
\end{table}
% Overall test NSE increased from 0.777 to 0.877, while RMSE decreased from 5.2 to 3.84 inches, an approximately $26\%$ reduction in prediction error. These improvements indicate that this approach more accurately captures the temporal evolution of SWE than climatological averages alone.

\begin{figure}[h]
    \centering
    \includegraphics[width=\linewidth]{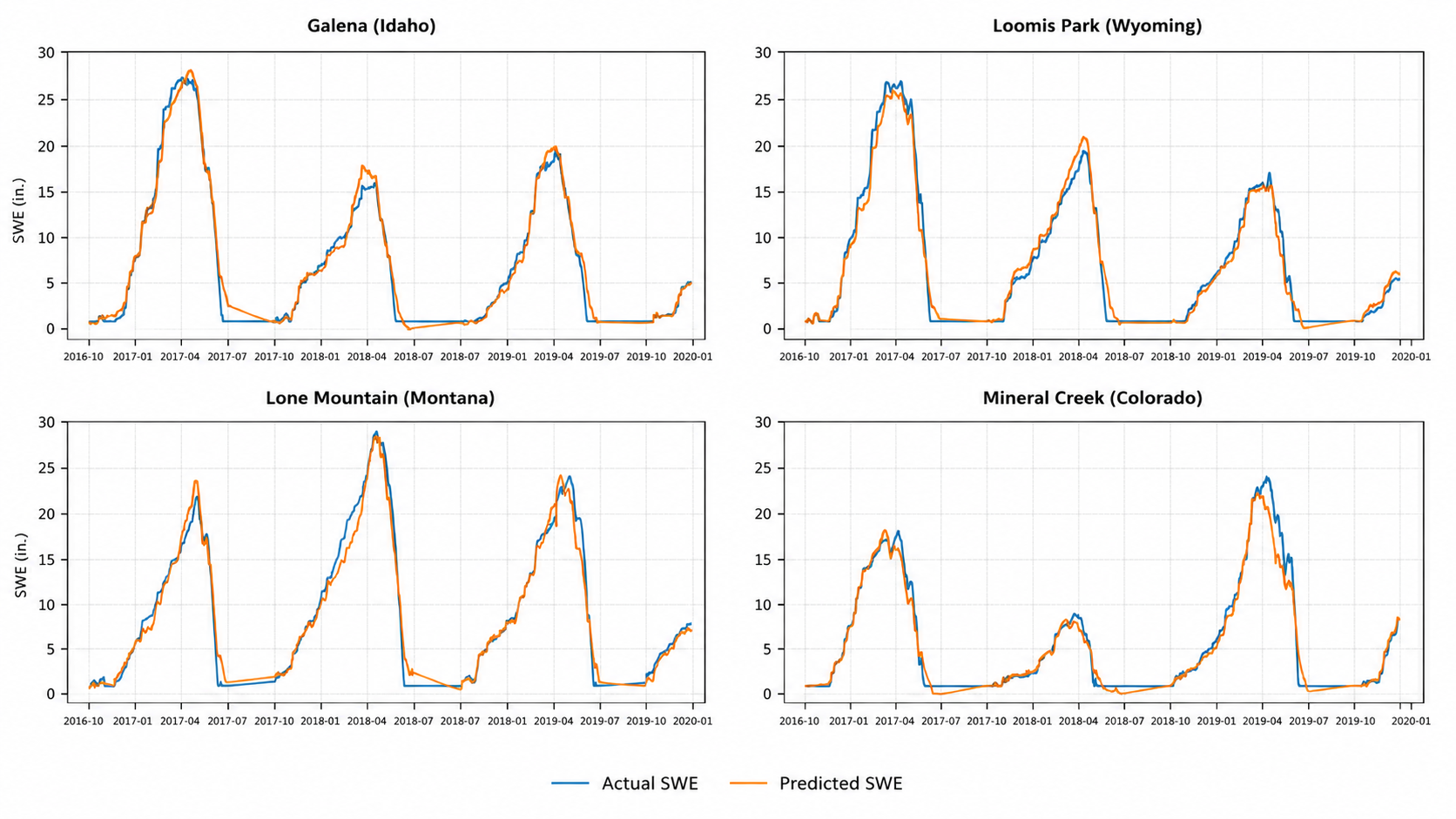}
    \caption{Observed (blue) and predicted (orange) SWE time series for four representative SNOTEL stations during the held-out test period. The spatiotemporal whitening framework accurately reproduces seasonal accumulation and melt dynamics across multiple snow years.}    \label{fig:timeseries}
\end{figure}
Figure~\ref{fig:timeseries} presents forecasts for four stations selected to represent a range of geographic regions and elevations across the western United States. We selected these stations to illustrate typical model behavior rather than to highlight the highest-performing locations. Across all four locations, our spatiotemporal predictions closely reproduce the seasonal evolution of SWE, accurately capturing both snow accumulation and spring melt over multiple snow years, although small discrepancies appear near peak accumulation and during rapid melt events. The model also captures interannual variability in peak SWE, indicating that it adapts to differences in snowpack magnitude.

Two features of our predictive model negatively impact NSE. Specifically, SWE is by definition non-negative, meaning predictions below zero are physically impossible. An interesting discussion also surrounds the model's occasional hesitance to approach zero: in some cases, instead of simply predicting zero after the melt season, predictions asymptotically approach zero. Both behaviors can increase prediction error near the end of the snow season, thereby reducing NSE even when overall SWE seasonality is captured well.

% , the predicted time series generally remain well aligned with the observed measurements. This visual representation results complement the quantitative improvements reported in Table~\ref{baseline-results-comparison} and demonstrate that our spatiotemporal whitening framework captures both the timing and magnitude of seasonal SWE dynamics across diverse hydrologic settings.

\subsection{Impact of Spatial Decorrelation}
To evaluate the contribution of the spatial whitening transformation, we compare results from training the LSTM on the decorrelated spatial data with an otherwise identical model trained directly on the original, spatially correlated observations. Both models share the same network architecture, hyperparameters, training procedure, and data splits. We thus attribute differences in predictive performance solely to the spatiotemporal whitening transformation.
\begin{figure}[h]
    \centering
    \includegraphics[width=\linewidth]{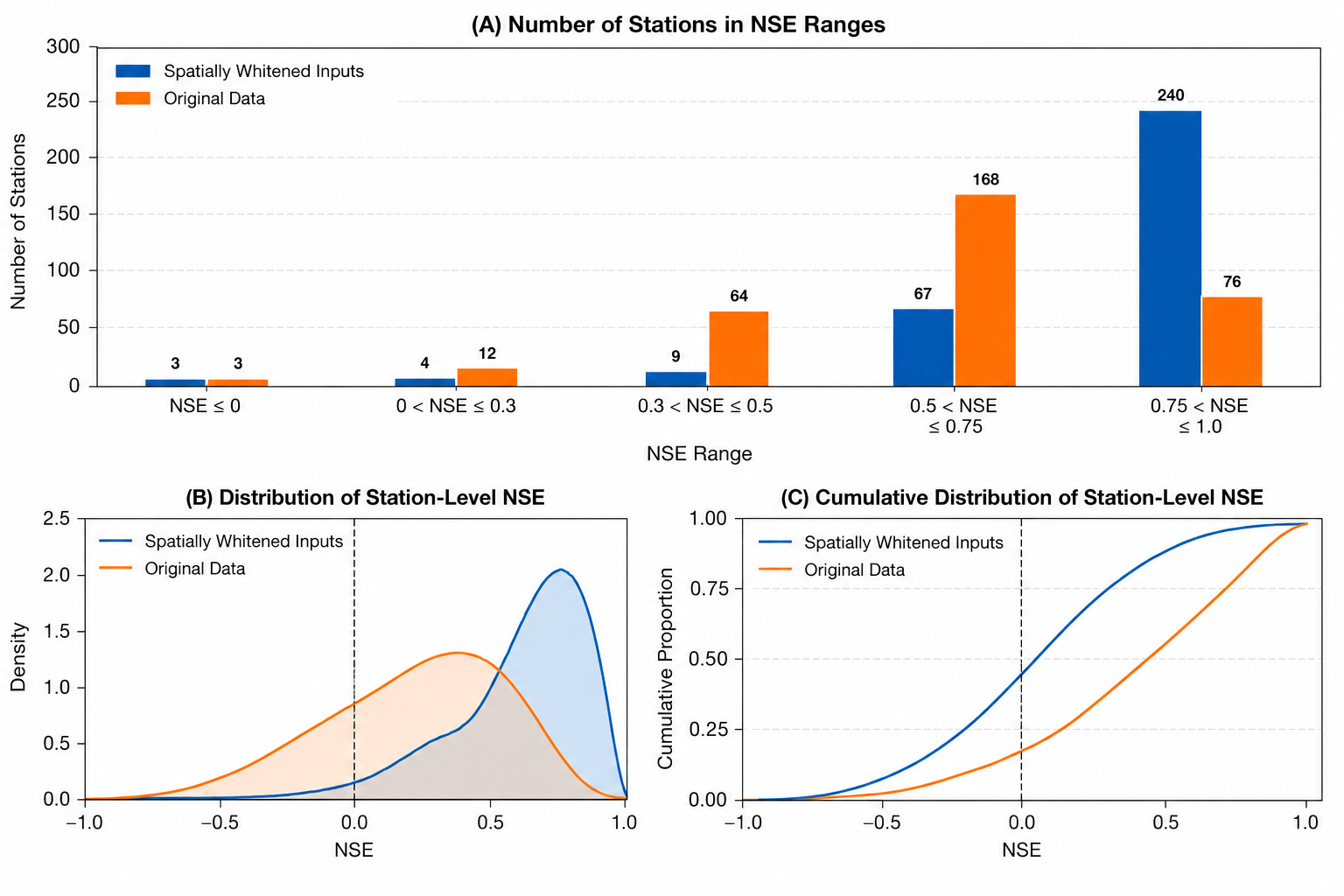}
    \caption{Comparison of station-level forecasting performance between the implemented spatially whitened framework and an otherwise identical LSTM trained on the original spatially correlated data. Panel (A) shows the number of stations within each NSE range, while Panels (B) and (C) summarize the distribution and cumulative distribution of station-level NSE values, respectively.}
    \label{fig:whitening-comparison}
\end{figure}

Figure~\ref{fig:whitening-comparison} demonstrates that spatial whitening substantially improves forecasting performance. %Our framework achieves an overall NSE of 0.877, compared with 0.708 for the LSTM trained on the original data, a nearly $24\%$ increase in forecasting accuracy. Likewise, RMSE decreases from 5.92 inches to 3.84 inches, representing an approximately $35\%$ reduction in prediction error.
The improvement extends beyond the aggregate performance metrics. As demonstrated by Figure~\ref{fig:whitening-comparison}, when trained on the original data, only 76 of the 323 stations achieve an NSE greater than 0.75. After applying the spatial whitening transformation, this number increases to 240 stations. %At the same time, the number of stations with moderate predictive performance decreases substantially.%, indicating that the spatiotemporal whitening transformation improves forecasting accuracy across much of the Western United States rather than only at a small subset of locations.

In short, by removing spatial dependence prior to model training, the whitening transformation reduces the burden of learning both spatial and temporal dependence simultaneously. This allows the LSTM to leverage its inductive bias toward sequential data while preserving the spatial covariance structure through the back-transformation described in Section~\ref{sec:spatial_whitening}.

\subsection{Geographic Variation in Forecast Accuracy}
\label{subsec:geovar}

\begin{figure}[h]
    \centering
    \includegraphics[width=0.5\linewidth]{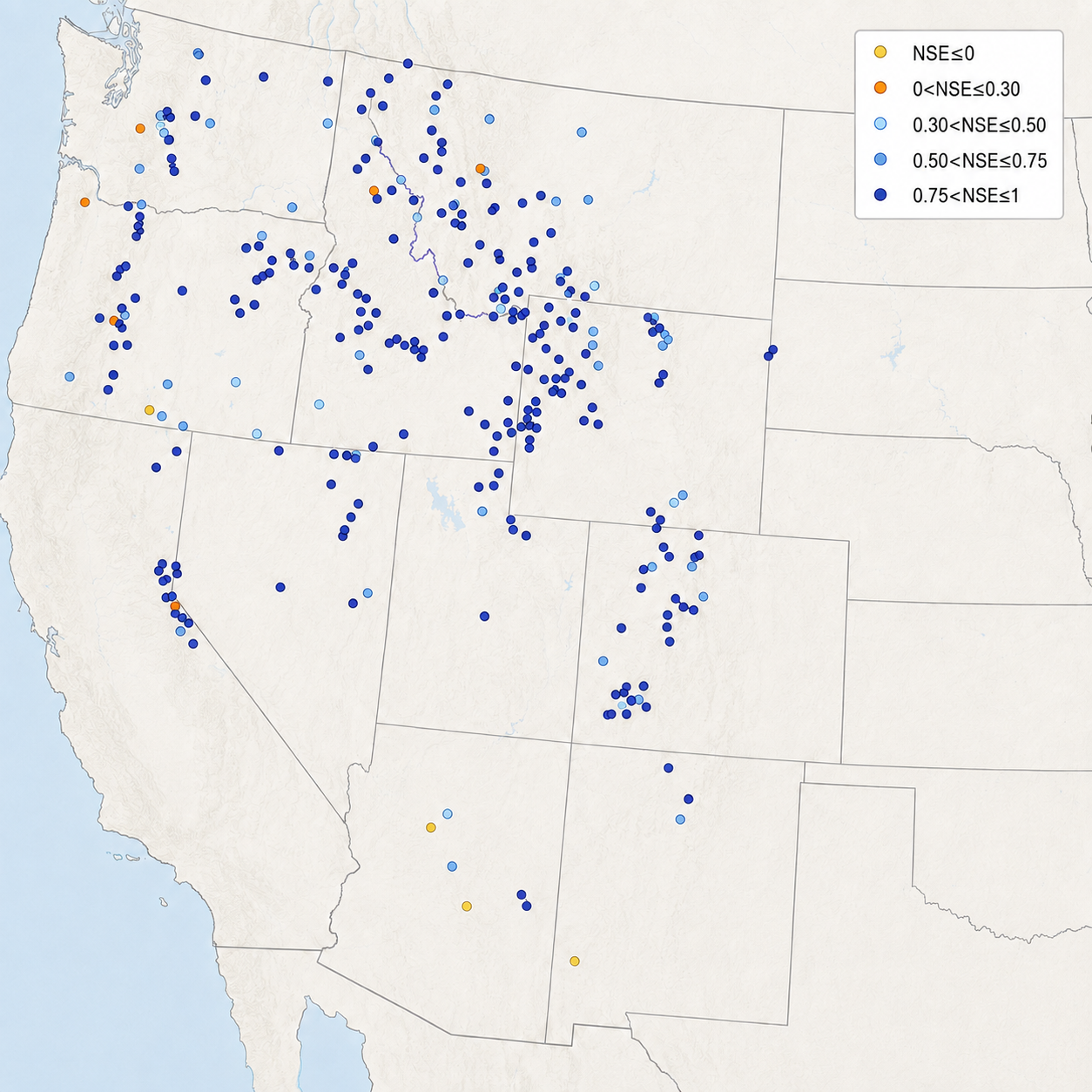}
    \caption{Individual SNOTEL Station NSE Results}
    \label{fig:station_map}
\end{figure}

Figure~\ref{fig:station_map} illustrates the spatial distribution of station-level NSE values across the 323 SNOTEL stations. Although forecasting accuracy varies geographically, our spatiotemporal framework achieves high predictive skill throughout much of the western United States.
%\FloatBarrier
Most stations achieve NSE values exceeding 0.75, particularly throughout the central Rocky Mountains and much of the Intermountain West. Lower NSE values occur primarily at a small number of stations in the Pacific Northwest and the southwestern United States, with a few isolated lower-performing stations appearing elsewhere in the domain. In general, it appears that stations with more persistent snowpacks elicit more accurate predictions, while stations with more ephemeral snowpacks are less predictable. In addition, regions with a lower spatial density of SNOTEL stations reduce the effectiveness of the spatial whitening transformation, contributing to diminished forecasting skill.

\subsubsection{A Failure Case: Signal Peak, NM}

To better understand the limitations of our spatiotemporal model, we examine Signal Peak, New Mexico, the most challenging station of the 323 analyzed for this work. 
\begin{table}[ht]
\centering
\caption{Summary Statistics for the Signal Peak SNOTEL station.}
\vspace{3mm}
\label{tab:signal_peak_summary}
\begin{tabular}{lr}
\hline
\textbf{Metric} & \textbf{Value} \\
\hline
NSE & $-6.30$ \\
Mean SWE & $0.903$ in \\
Dataset Mean SWE & $7.92$ in \\
Mean Distance to 10 Nearest Neighbors & $362$ km \\
Neighbor Distance Rank & $323/323$ \\
\hline
\end{tabular}
\end{table}
%\FloatBarrier
\begin{figure}[h]
    \centering
    \includegraphics[width=.75\linewidth]{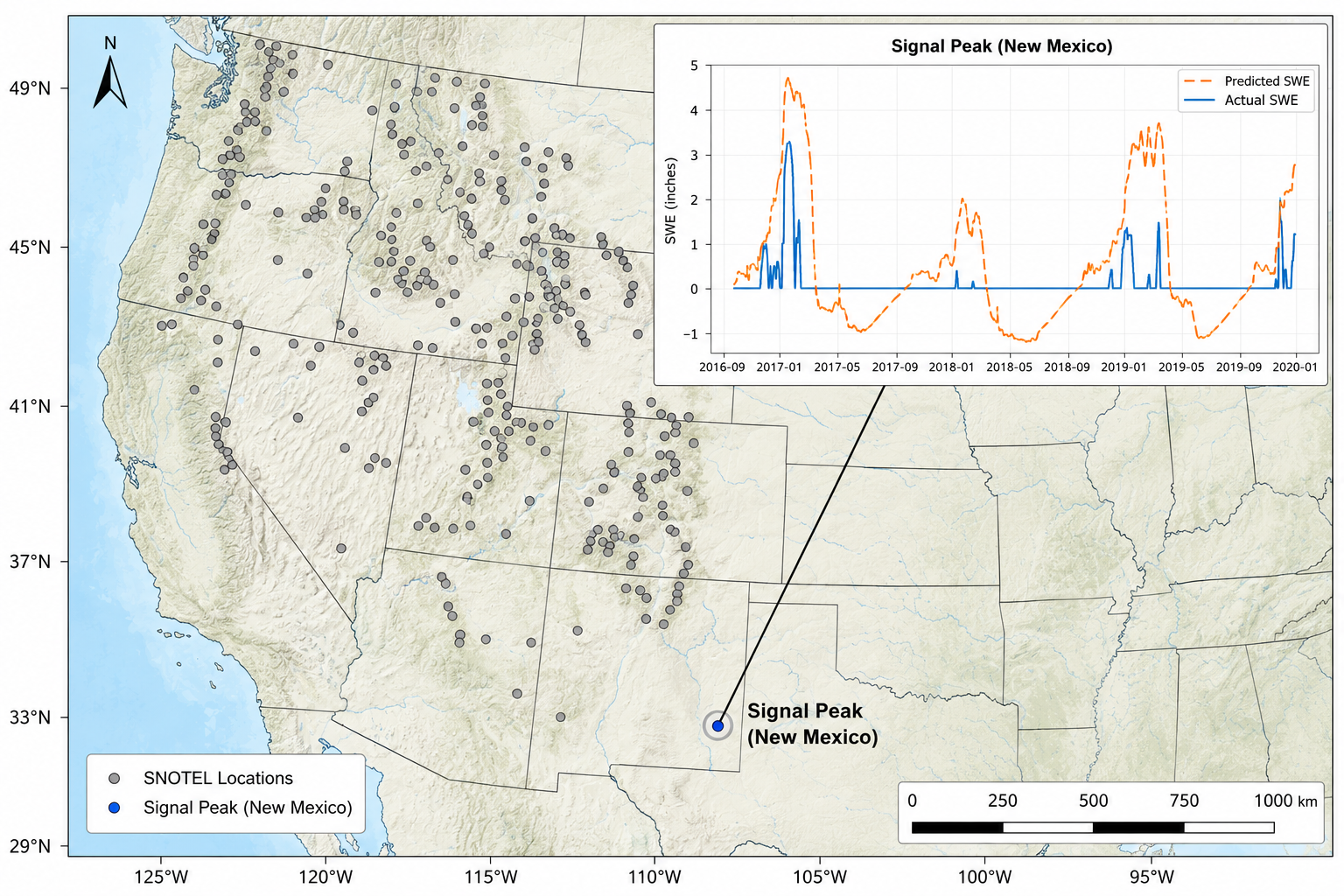}
    \caption{Map of 323 SNOTEL stations with inset containing observed (blue) and predicted (orange) SWE time series for Signal Peak, NM.}
    \label{fig:signalpeak}
\end{figure}
\FloatBarrier
Signal Peak has three characteristics that distinguish it from nearly all other SNOTEL stations and make it an unusually challenging prediction problem. First, it is the most spatially isolated station in the dataset, with a mean distance of 362 km (225 mi) to its ten nearest neighbors, reducing the local spatial information available to the whitening transformation. Second, it exhibits an average SWE of only 0.903 inches, compared with a network-wide average of 7.92 inches, yielding a substantially weaker seasonal signal for temporal learning. Third, as shown by the inset of Figure~\ref{fig:signalpeak}, snow accumulation is both shallow and highly intermittent. 

However, although Signal Peak represents the most challenging station studied, its combination of extreme spatial isolation and exceptionally low seasonal snowpack is uncommon within the SNOTEL network. This failure case therefore illustrates how the spatiotemporal prediction method can break down when hindered by both a lack of spatial information and a weak seasonal SWE pattern.

\subsection{Conformal Prediction Interval Analysis}
\label{sec:results_conformal_prediction}

We next assess the uncertainty estimates produced by the split conformal prediction procedure described in Section~\ref{sec:methodology_conformal_prediction}. Because we perform calibration separately for each day of the snow year, the resulting conformal half-width $q_{1-\alpha}(t)$ varies seasonally but remains common across stations for a fixed snow day. We therefore examine the variation in these snow-day-specific quantiles throughout the snow year together with the empirical coverage obtained on the held-out test snow years.

\begin{figure}[h]
\centering
\includegraphics[width=.75\linewidth]{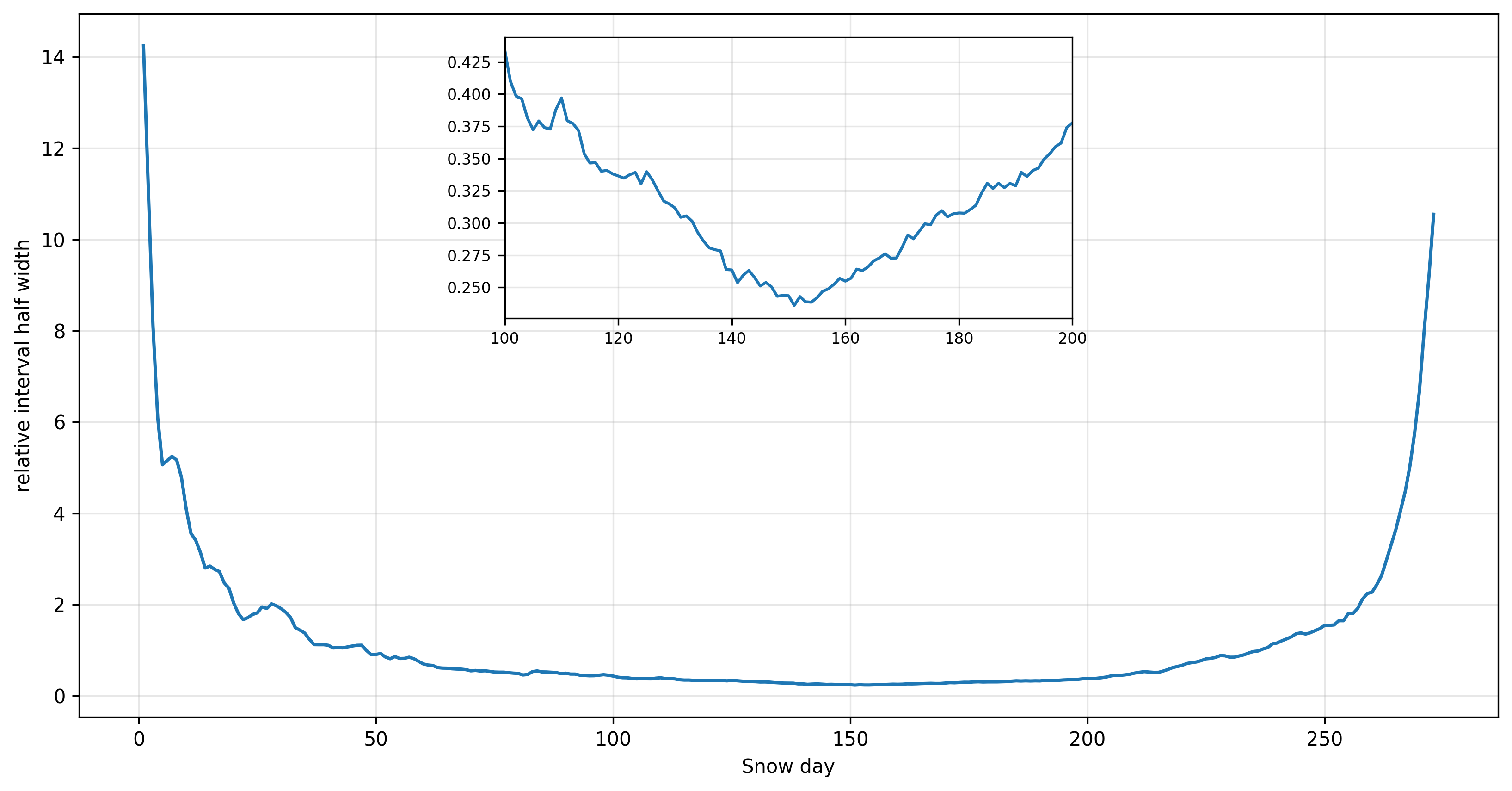}
\caption{Relative conformal half-width by snow day, computed by dividing the conformal half-width $q_{1-\alpha}(t)$ by the mean SWE for each specific day.}
\label{fig:cp_quantile_by_day}
\end{figure}
%\FloatBarrier
Figure~\ref{fig:cp_quantile_by_day} shows how the {\em relative} conformal interval half-width---i.e., $q_{1-\alpha}(t)$, divided by the mean observed SWE for day $t$---varies throughout the snow year. % Smaller values indicate that the conformal interval is narrow relative to the mean SWE, whereas larger values indicate greater uncertainty relative to the prevailing snowpack magnitude. 
Because the calibration procedure pools residuals across all stations for each snow day, the temporal variation in Figure~\ref{fig:cp_quantile_by_day} reflects seasonal changes in forecasting difficulty rather than spatial differences among individual stations.

Across the full snow year, the relative conformal half-width ranged from $0.24$ (in other words, the size of the interval is $24\%$ of the mean) to $14.24$, with a median of $0.52$ and a mean of $1.20$. The largest values occurred near the beginning and end of the snow year, when mean SWE approached zero and rapid accumulation and melt cycles occurred, driving interval widths to increase relative to the average snowpack. In contrast, the relative half-width remained substantially smaller during the central portion of the snow year, when mean SWE was larger and snowpack changes occurred more gradually. The inset highlights snow days 100--200 and shows that the relative half-width decreased toward the middle of this period before increasing again, indicating that relative uncertainty continued to vary systematically even during the primary accumulation and peak-snowpack portion of the season. 

\begin{figure}[h]
\centering
\includegraphics[width=\linewidth]{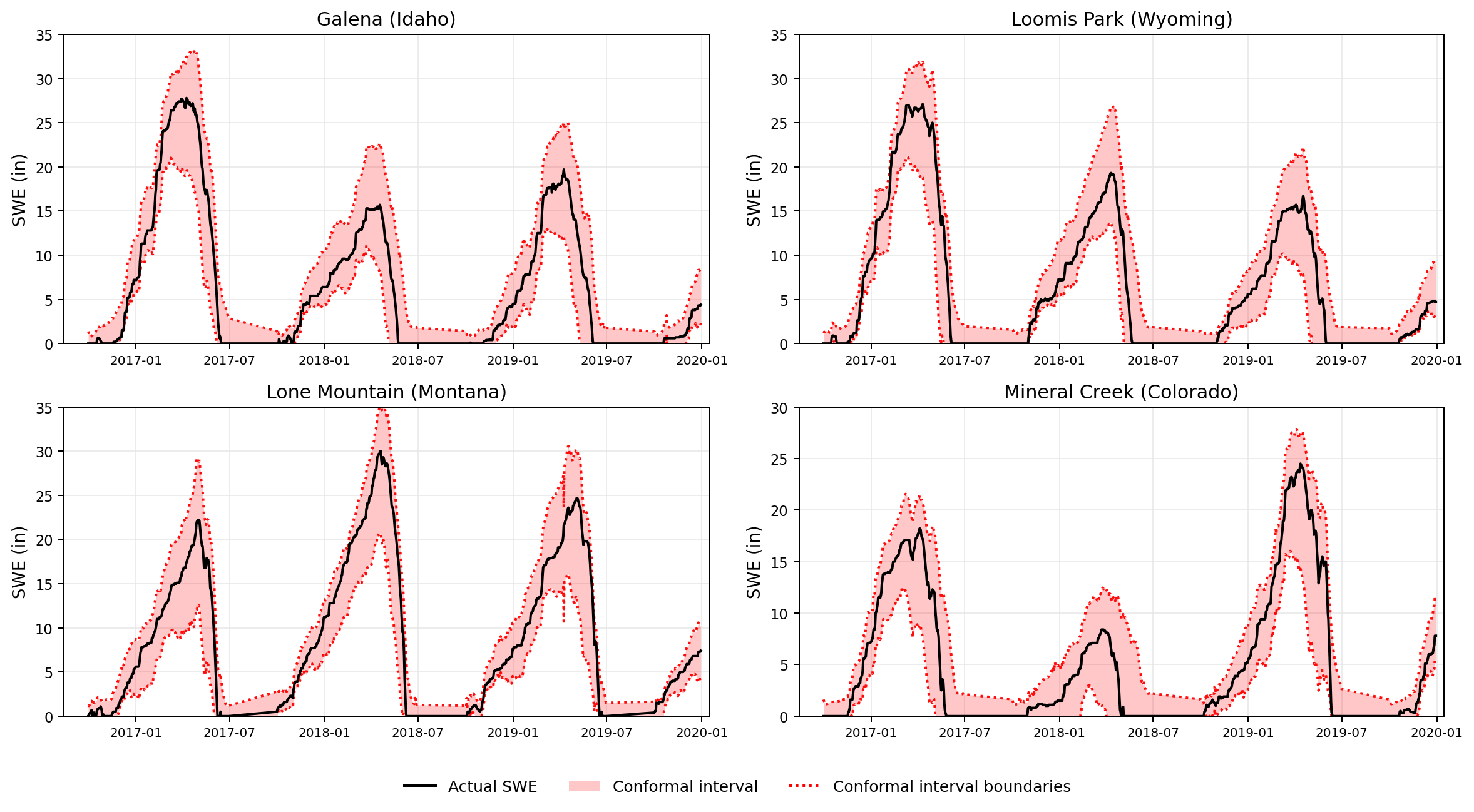}
\caption{Nominal 90\% conformal prediction intervals for four representative SNOTEL stations during the held-out test period. The black line shows the observed SWE, the shaded region shows the conformal prediction interval, and the red dotted lines show the lower and upper interval boundaries.}
\label{fig:cp_intervals_selected_stations}
\end{figure}
%\FloatBarrier

Figure~\ref{fig:cp_intervals_selected_stations} illustrates the conformal prediction intervals at four representative stations (the same sites as in Figure~\ref{fig:timeseries}); the intervals generally follow the seasonal accumulation and melt cycles observed at each location. They widen during periods of greater snow accumulation, when the magnitude and variability of the prediction errors tend to increase, and narrow near the beginning and end of each snow season. We note that at these four stations, the observed SWE remains within the conformal interval for almost all of the test period.

% , only crossing an interval boundary 

% The intervals capture both the gradual accumulation phase and the rapid decrease in SWE during spring melt. However, the observed series occasionally approaches or crosses an interval boundary, particularly near seasonal peaks and during rapid melt events. These departures help explain why the overall PICP falls below the nominal 90\% level.

\begin{figure}[h]
\centering
\includegraphics[width=\linewidth]{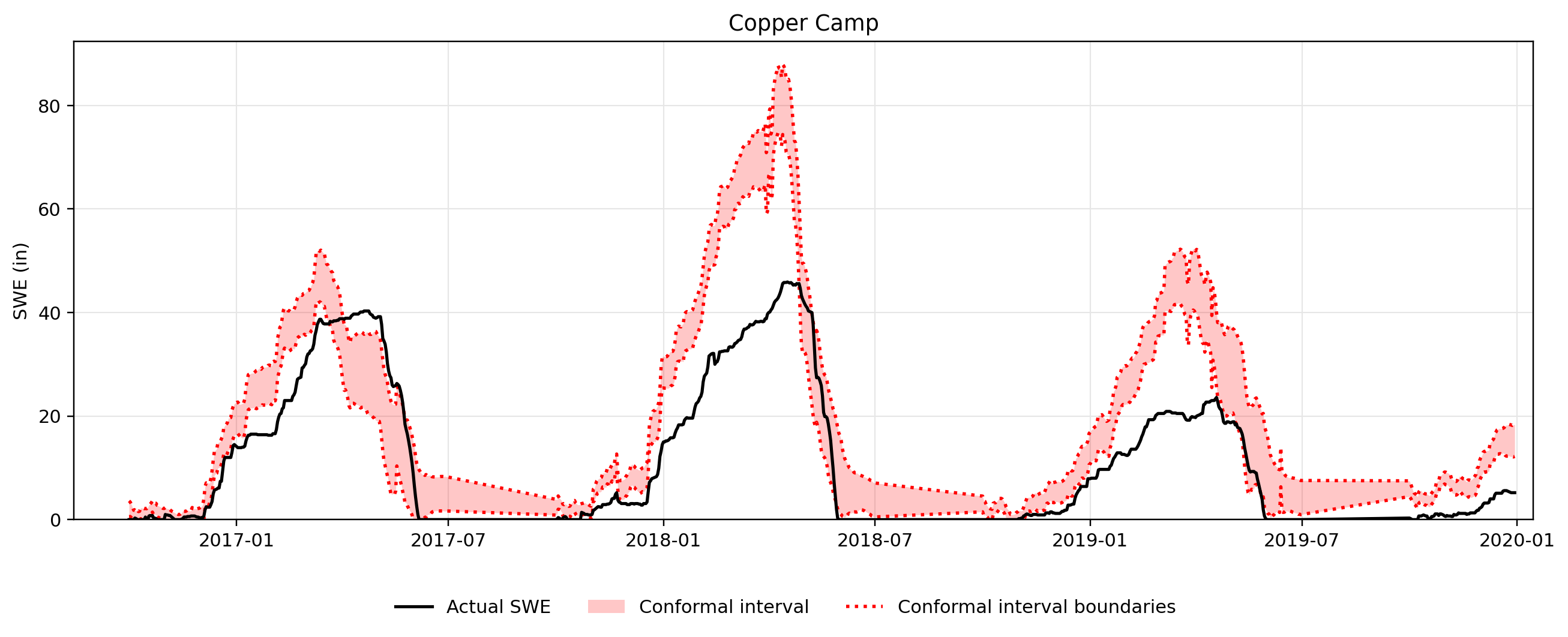}
\caption{Nominal 90\% conformal prediction interval for Copper Camp during the held-out test period.}
\label{fig:cp_copper_camp}
\end{figure}

However, Figure~\ref{fig:cp_copper_camp} demonstrates the shortcomings with conformal prediction in this application; that is, it cannot correct systematic errors in the underlying forecasting model. At the Copper Camp (Montana) SNOTEL site, for example, the model substantially overpredicts SWE during the majority of the test years. Because the conformal interval is centered on the model prediction, this bias shifts the entire interval upward and produces upper bounds that greatly exceed the observed SWE. Consequently, stations with significant prediction errors can retain poorly centered intervals even when the broader calibration procedure provides reasonable marginal coverage.

%Furthermore, we note this example also explains a lower PICP and provides important context on conformal prediction interval quality. Conformal prediction sets the interval width using the calibration residuals, but it does not alter the point prediction or remove station-specific bias. 

Finally, we calculate PICP (see Section~\ref{sec:PICP}) on the held-out test snow years and present the results in Table~\ref{tab:cp_coverage_summary}. Across the held-out test observations, the snow-day-specific conformal prediction intervals achieved an overall PICP of 85.89\%, below the nominal target of 90\%. Snow-day-level coverage ranged from 76.55\% to 92.95\%, with a median of 86.38\%, indicating that the intervals were not uniformly calibrated throughout the snow season. Two potential explanations for this undercoverage exist: pooling residuals across all stations for a fixed snow day may obscure substantial differences in station-level forecasting uncertainty, while differences between calibration and testing snow years may introduce interannual shifts in the error distribution.

\begin{table}[h]
\centering
\caption{PICP results for snow-day-specific conformal prediction intervals.}
\vspace{3mm}
\label{tab:cp_coverage_summary}
\begin{tabular}{lc}
\hline
\textbf{Metric} & \textbf{Value} \\
\hline
Overall coverage & 85.89\% \\
Mean snow-day coverage & 85.73\% \\
Median snow-day coverage & 86.38\% \\
Minimum snow-day coverage & 76.55\% \\
Maximum snow-day coverage & 92.95\% \\
Mean interval half-width & 3.89 in. \\
Median interval half-width & 3.43 in. \\
\hline
\end{tabular}
\end{table}
%\FloatBarrier

Accordingly, we  interpret the snow-day-specific conformal procedure as an empirical approach to uncertainty quantification in hydrologic applications rather than as an application which establishes the exact finite-sample coverage guarantee. Other approaches, such as station-specific split conformal prediction, may yield interpretable results but lack the formal theoretical guarantees resulting from the exchangeability requirement.

\section{Conclusion}
%\subsection{Discussion}
%This study highlighted that a GP-based spatial decorrelation framework can sufficiently separate spatial and temporal dependence to improve spatiotemporal SWE forecasting.
Building upon the spatial whitening methodology of \citet{cite-heaton}, we combined spatial decorrelation with an LSTM forecasting model and distribution-free conformal prediction to produce a forecasting framework that delivers both accurate point predictions and measures of predictive uncertainty for sequential SWE data, with results demonstrating the predictive accuracy of this framework. Relative to both a climatological baseline and an otherwise identical LSTM trained on the original spatially correlated observations, the model substantially improved forecasting accuracy across the 323 SNOTEL stations. We note that our model also improved on modern predictive frameworks using advanced DL architectures, such as \citet{cite-Thapa}. These improvements occurred throughout much of the western United States, with a dramatic increase in NSE occurring at nearly all stations after applying the spatial whitening transformation. 

We attribute these improvements to the separation of spatial and temporal learning. Rather than requiring a single model to simultaneously learn spatial covariance and temporal signals, we delegate these tasks to methods that are more naturally suited to each individual problem. The GP-based whitening transformation accounts for local spatial dependence before training, while the LSTM models the sequential evolution of SWE throughout each snow year. The decorrelation reduces the complexity of the learning task and provides a principled alternative to increasingly complex end-to-end architectures that attempt to learn both dependence structures simultaneously.

Beyond improving point forecasting performance, this work also provides uncertainty quantification through conformal prediction. This extends the impacts of the spatiotemporal forecasting model, as many water-resource management decisions require explicit characterization of predictive uncertainty. %For instance, flood mitigation strategies, drought planning, and seasonal water allocation decisions all depend on understanding the range of plausible future SWE values rather than a single predictive value. 
By combining conformal prediction with the spatial whitening-based forecasting framework, we construct prediction intervals without requiring additional distributional assumptions regarding model residuals.

While strong in its predictive accuracy, our approach has certain limitations. First, this project is specifically aimed at point prediction, meaning that while future methods may emerge for spatiotemporally complete SWE forecasting using this approach, we predicted only at these 323 specific points. As a result, the model is best interpreted as a nowcasting tool rather than a fully forward-looking forecasting system. Second, we assume a stationary covariance structure and approximate spatial dependence using a fixed nearest-neighbor representation. This may not fully capture regional differences in snow processes or evolving climatic conditions. Finally, the transformation proposed by \citet{cite-heaton} is derived under a GP framework in which the spatial process is decomposed into a linear mean component and a zero-mean spatial residual. The proposed spatiotemporal approach in this work, however, models the conditional mean using a nonlinear neural network. While the whitening transformation is applied in the same manner, there is currently no formal theoretical guarantee that the decorrelation procedure preserves its properties under an arbitrary nonlinear mean model; the approach adopted here should thus be viewed only as a practical extension of the decorrelation method.

A natural extension is to develop a spatiotemporally complete SWE prediction system that integrates in situ observations with reanalysis data, satellite products, and other remotely sensed information. Likewise, relaxing the assumption of stationary spatial covariance through adaptive or nonstationary covariance models may further improve forecasting performance in regions with complex snow dynamics. Finally, leveraging attention-head mechanisms \citet{cite-attention}, U-Nets \citet{cite-Unets,sikorski2025latticevision}, and other convolutional and recurrent neural networks in lieu of LSTMs may offer interesting methods for predictive accuracy comparisons.

% \subsection{Closing Thoughts}
% By explicitly separating spatial and temporal dependence, this work demonstrates that applying a statistically sound spatial structure through the whitening transformation allows relatively simple temporal models to achieve strong predictive skill. The resulting framework is both computationally efficient and interpretable, providing a principled alternative to attention-based and graph-based methods for environmental time series prediction.

\subsection*{Acknowledgments and Funding Information}
CF's graduate education was funded and supported by the U.S. Coast Guard. SB's work has been partially supported by NSF CMMI 2210840 and NSF DMS 2514857. 

\subsection*{Data Availability Statement}

The data that support the findings and all code responsible for implementing the spatiotemporal whitening predictive approach are openly available and can be accessed at \url{https://github.com/mines-opt-ml/autoswe/}. %The repository contains the necessary code for decorrelating spatial data, training the subsequent LSTM model, evaluating the test set, and applying split conformal prediction for uncertainty quantification. The repository also includes a configuration file which can adapt to different hyperparameters or data sources at the request of the user. 

%\newpage
\bibliographystyle{unsrtnat}
\bibliography{citations}
\appendix
\section{A Note on Gaussianity}
\label{app:Gaussianity}
The spatial whitening transformation relies on a GP model for the spatial covariance structure of the data; that is, it assumes that the spatial residual process is approximately jointly Gaussian after accounting for the mean structure.

We investigated Gaussianity at several levels. Normality tests (e.g. Lilliefors \citet{cite-Lilliefors1967, cite-VanSoest1967}, D'Agostono-Pearson \citet{cite-DAgostinoPearson1973}) applied to raw SWE values pooled across stations and dates strongly rejected a Gaussian distribution. Similar tests applied separately across time for each station and across stations for each date also rejected normality in nearly all cases. These results were expected, as raw SWE observations are nonnegative, strongly seasonal, frequently equal to zero, and drawn from stations with substantially different elevations, climates, and accumulation patterns. Pooling these observations therefore combines multiple distinct distributions and does not directly test the Gaussian assumption underlying the spatial process, though it offers interesting discussion on underlying normality in spatial observations. For instance, \citet{cite-Liljestrand2025} observed an approximately Gaussian snow-depth distribution within a subalpine basin, but with caveats: snow-based distributions commonly exhibit skewness during accumulation and melt periods, in wind-affected terrain, and where snow-free areas occur. We thus should proceed with caution when assigning a single fixed probability density to snow-related data \citet{cite-Ohara2024}. 

%For instance, many statistical methods for SWE estimation and prediction represent spatial variability through a Gaussian framework \citet{cite-Aerenson2025}. However, the use of a GP does not require the observed SWE values themselves to follow a Gaussian marginal distribution; empirical snow distributions may exhibit substantial non-Gaussianity, with the degree of departure from Gaussianity varying across spatial regions \citet{cite-Ohara2024}.

We next examined prediction residuals on both the original and spatially decorrelated scales. Residuals on the original SWE scale showed substantial positive skewness and heavy tails. After spatial decorrelation, the residual distribution became nearly symmetric, with pooled skewness decreasing from approximately $2.29$ on the original scale to $0.09$ on the decorrelated scale. However, the decorrelated residuals retained substantial excess kurtosis, and formal normality tests continued to mostly reject exact Gaussianity. Thus, while the spatial transformation considerably reduced asymmetry, it did not produce exactly Gaussian residuals.

Therefore, as with other aspects of our spatiotemporal application of the whitening process proposed by \citet{cite-heaton}, we interpret the Gaussianity assumption as a working approximation rather than an exact description of the physical SWE process. 

\section{The Cholesky Decomposition and Spatial Whitening}
\label{app:cholesky}

The Cholesky decomposition provides a numerically stable method for solving linear systems involving real, symmetric, positive-definite matrices. For any such matrix $Z\in\mathbb{R}^{p\times p}$, the decomposition takes the form
\begin{equation}
    Z = LL^{\top},
    \label{eq:cholesky}
\end{equation}
where $L$ is a lower-triangular matrix. Suppose that we wish to solve $Z\bm{x}=\bm{b}$. After substituting \eqref{eq:cholesky}, we obtain $LL^{\top}\bm{x}=\bm{b}$. The solution can then be found using the two triangular systems
\begin{align}
    L\bm{u} &= \bm{b},
    \label{eq:cholesky_forward}\\
    L^{\top}\bm{x} &= \bm{u}.
    \label{eq:cholesky_backward}
\end{align}
We can solve equations~\eqref{eq:cholesky_forward} and \eqref{eq:cholesky_backward} through forward and backward substitution, respectively. This is in general the fastest way to solve a linear system, with a total operation count of $\frac{1}{3}m^3 + O(m^2)$ \citet{trefethen2022numerical}. The Cholesky decomposition also plays an important role in GP computation. For example, if a covariance matrix satisfies $\Sigma = LL^{\top}$ and $ \bm{z}\sim\mathcal{N}(\bm{0},I)$ then $L\bm{z}\sim\mathcal{N}(\bm{0},\Sigma)$. 
%More directly relevant to this work, Cholesky factorization allows the local covariance systems arising from the Vecchia approximation to be solved without explicitly forming covariance-matrix inverses.

\subsection{Computation of the Local Kriging Weights}

Recall that the Kriging weights for location $s_i$ are defined by
\begin{equation}
    \bm{w}_i^{\top}
    =
    R(i,\mathcal{C}_i)
    R(\mathcal{C}_i,\mathcal{C}_i)^{-1},
    \label{eq:app_kriging_weights}
\end{equation}
where $\mathcal{C}_i$ denotes the conditioning set associated with location $s_i$. Under the Vecchia construction, $\mathcal{C}_i$ contains at most $M$ predecessor locations according to the selected spatial ordering, so that $|\mathcal{C}_i| = \min\{M,i-1\}$.
However, recall that our implementation does not impose an ordering and instead defines $\mathcal{C}_i$ using the $M$ nearest spatial locations, so that $|\mathcal{C}_i|=M$.

Rather than evaluating the inverse in \eqref{eq:app_kriging_weights}, we transpose the expression and solve the equivalent linear system
\begin{equation}
    R(\mathcal{C}_i,\mathcal{C}_i)\bm{w}_i
    =
    R(\mathcal{C}_i,i).
    \label{eq:local_weight_system}
\end{equation}
Because the local correlation matrix is symmetric and positive definite, it admits the Cholesky factorization $R(\mathcal{C}_i,\mathcal{C}_i) = L_iL_i^{\top}$. Using the method described in Section~\ref{app:cholesky}, the Kriging weights are obtained by solving $L_i\bm{u}_i$ and then $L_i^{\top}\bm{w}_i$.
% \begin{align}
%     L_i\bm{u}_i
%     &=
%     R(\mathcal{C}_i,i),
%     \label{eq:local_forward}\\
%     L_i^{\top}\bm{w}_i
%     &=
%     \bm{u}_i.
%     \label{eq:local_backward}
% \end{align}

Importantly, this procedure does not require a Cholesky decomposition of the full $N\times N$ covariance matrix $\Sigma$ (or equivalently of the full correlation matrix $R$). Instead, it performs a collection of local factorizations involving matrices of dimension at most $M\times M$. A full GP factorization generally requires computational cost on the order of $O(N^3)$, whereas the collection of local Vecchia factorizations have a complexity bounded above by $O(NM^3)$. Because $M\ll N$, the local construction substantially reduces the cost of obtaining the spatial transformation.

The nugget proportion (see \eqref{eq:Sigma_defn}) contributes to the numerical stability of these calculations by improving the conditioning of the
$R(\mathcal{C}_i,\mathcal{C}_i)$ and helping to preserve positive definiteness during Cholesky factorization.

For the spatiotemporal setting considered in this work, we assume the station locations do not change over time. Thus, the station ordering, conditioning sets, neighborhood size, and covariance parameters remain fixed after specifying the spatial model. It follows that $\mathcal{C}_i, \bm{w}_i, \text{ and } \nu_i$ are time independent.

\section{Further details on LSTMs}
\label{app:lstm}
We implement the forecasting function $\widehat f$ using a LSTM neural network. Unlike a standard feedforward neural network, the LSTM processes sequential inputs recursively and maintains an internal memory state that enables the model to learn both short-term and long-term temporal dependencies. 

More specifically, let $\bm{x}_t$ denote the input vector at time step $t$, $\bm{h}_t$ the hidden state, and $\bm{c}_t$ the cell state. The LSTM updates these states according to
\[
\bm{f}_t
=
\sigma
\left(
\bm{W}_f \bm{x}_t
+
\bm{U}_f \bm{h}_{t-1}
+
\bm{b}_f
\right),
\]
\[
\bm{i}_t
=
\sigma
\left(
\bm{W}_i \bm{x}_t
+
\bm{U}_i \bm{h}_{t-1}
+
\bm{b}_i
\right),
\]
\[
\widetilde{\bm{c}}_t
=
\tanh
\left(
\bm{W}_c \bm{x}_t
+
\bm{U}_c \bm{h}_{t-1}
+
\bm{b}_c
\right),
\]
\[
\bm{c}_t
=
\bm{f}_t \odot \bm{c}_{t-1}
+
\bm{i}_t \odot \widetilde{\bm{c}}_t,
\]
\[
\bm{o}_t
=
\sigma
\left(
\bm{W}_o \bm{x}_t
+
\bm{U}_o \bm{h}_{t-1}
+
\bm{b}_o
\right),
\]
and
\[
\bm{h}_t
=
\bm{o}_t \odot \tanh(\bm{c}_t),
\]

where $\sigma(\cdot)$ denotes the logistic sigmoid function, $\tanh(\cdot)$ denotes the hyperbolic tangent function, and $\odot$ denotes elementwise multiplication. The matrices $\bm{W}_f$, $\bm{W}_i$, $\bm{W}_c$, $\bm{W}_o$, $\bm{U}_f$, $\bm{U}_i$, $\bm{U}_c$, and $\bm{U}_o$, together with the bias vectors $\bm{b}_f$, $\bm{b}_i$, $\bm{b}_c$, and $\bm{b}_o$, constitute the trainable parameters of the network. 

We subsequently define
\[
\theta
=
\{
\bm{W}_f,\bm{W}_i,\bm{W}_c,\bm{W}_o,
\bm{U}_f,\bm{U}_i,\bm{U}_c,\bm{U}_o,
\bm{b}_f,\bm{b}_i,\bm{b}_c,\bm{b}_o,
\bm{W}_y,\bm{b}_y
\}
\]
as the collection of all trainable parameters and note that at each time step, the corresponding hidden state is mapped to a scalar prediction through a fully connected output layer,
\[
    \widehat{\widetilde{y}}_{i,a,t}
    =
    \bm{W}_y\bm{h}_t + \bm{b}_y,
    \qquad t=1,\ldots,T.
\]
Applying this output layer at each time step produces the sequence-to-sequence prediction defined in Section~\ref{sec:spatiotemporal_training}. 

To compute parameter counts, we first specify that the dynamic forcing vector contains seven input variables, with two station-specific variables---the station-specific mean and standard deviation of SWE---appended prior to the LSTM, giving a total input dimension of $F=9$. For clarity, we note the seven inputs are the features outlined in Section~\ref{subsec:features}: maximum/minimum/observed temperature, precipitation, thermal brightness 19/37 GHz, and thermal brightness difference. We use a hidden-state dimension of $H=256$ and four recurrent layers. For the first recurrent layer, each input-weight matrix $\bm{W}_f,\bm{W}_i,\bm{W}_c,\bm{W}_o$ has dimension $256\times9$, while each recurrent-weight matrix $\bm{U}_f,\bm{U}_i,\bm{U}_c,\bm{U}_o$ has dimension $256\times256$. For each of the remaining three recurrent layers, both the input- and recurrent-weight matrices have dimension $256\times256$. The fully connected output layer maps the $256$-dimensional hidden state to a scalar SWE prediction. In total, the implemented network therefore contains approximately $1,852,674$ trainable parameters.

\end{document}